%% file: main.tex
\documentclass{article}
\pdfoutput=1
\usepackage{microtype}
\usepackage{graphicx}
\usepackage{subcaption}
\usepackage{booktabs} %
\usepackage{hyperref}
\usepackage{bm}

\usepackage[accepted]{icml2021}

\usepackage[utf8]{inputenc} %
\usepackage[T1]{fontenc}    %
\usepackage{xcolor}

\definecolor{mydarkblue}{rgb}{0,0.08,0.45}

\usepackage{array}
\usepackage{url}            %
\usepackage{amsfonts}       %
\usepackage{nicefrac}       %
\usepackage{microtype}      %

\usepackage{amsmath}
\usepackage{multirow}
\usepackage{color}
\usepackage{bm}
\usepackage{amsthm}

\usepackage{stmaryrd}
\usepackage{grffile}%

\input{notation}

\icmltitlerunning{Implicit rate-constrained optimization of non-decomposable objectives}

\begin{document}
\newcommand{\proposed}{ICO}

\twocolumn[
\icmltitle{Implicit Rate-Constrained Optimization of Non-decomposable Objectives}

\begin{icmlauthorlist}
\icmlauthor{Abhishek Kumar}{goo}
\icmlauthor{Harikrishna Narasimhan}{goo}
\icmlauthor{Andrew Cotter}{goo}
\end{icmlauthorlist}

\icmlaffiliation{goo}{Google Research}

\icmlcorrespondingauthor{Abhishek Kumar}{abhishk@google.com}

\vskip 0.3in
]

\printAffiliationsAndNotice{}  %

\input{abstract.tex}

\input{intro}

\input{formulation}

\input{method}

\input{related}
\input{experiments}
\input{discussion}

\bibliography{ml}
\bibliographystyle{icml2021}

\clearpage
\appendix
\input{supplementary/appendix}

\end{document}

%% file: notation.tex
\newcommand{\punt}[1]{}

\newtheorem{exmp}{Example}

\newtheorem{theorem}{Theorem}

\newtheorem{prop}{Proposition}

\def\argmax{\mathop{\rm arg\,max}}

\newcommand{\reals}{\mathbb{R}}

\def\argmax{\mathop{\rm arg\,max}}

\newcommand{\bq}{\begin{equation}}
\newcommand{\eq}{\end{equation}}
\newcommand{\ba}{\begin{eqnarray}}
\newcommand{\ea}{\end{eqnarray}}

\def\R{{\reals}}

\newcommand{\remove}[1]{}

\newcommand{\0}{\bm{0}}

\newcommand{\ie}{\textit{i}.\textit{e}.}
\newcommand{\eg}{\textit{e}.\textit{g}.}
\newcommand{\etc}{\textit{etc}.}

\newcommand{\tf}{\tilde{f}}
\newcommand{\tg}{\tilde{g}}
\renewcommand{\th}{\tilde{h}}

\renewcommand{\P}{\mathbb{P}}
\newcommand{\E}{\mathbb{E}}

\renewcommand{\>}{{\rightarrow}}
\newcommand{\1}{\mathbf{1}}
\renewcommand{\prec}{\textup{\textrm{precision}}}
\newcommand{\rec}{\textup{\textrm{recall}}}
\newcommand{\tp}{\textup{\textrm{TP}}}
\newcommand{\tn}{\textup{\textrm{TN}}}
\newcommand{\fp}{\textup{\textrm{FP}}}
\newcommand{\fn}{\textup{\textrm{FN}}}
\newcommand{\tpr}{\textup{\textrm{TPR}}}
\newcommand{\fpr}{\textup{\textrm{FPR}}}
\newcommand{\fnr}{\textup{\textrm{FNR}}}

%% file: abstract.tex
\begin{abstract}
We consider a popular family of constrained optimization problems arising in machine learning that involve optimizing a non-decomposable evaluation metric with a certain thresholded form, while constraining another metric of interest. Examples of such problems include optimizing the false negative rate at a fixed false positive rate, optimizing precision at a fixed recall, optimizing the area under the precision-recall or ROC curves, etc. Our key idea is to formulate a rate-constrained optimization that expresses the threshold parameter as a function of the model parameters via the Implicit Function theorem. We show how the resulting optimization problem can be solved using standard gradient based methods. Experiments on benchmark datasets demonstrate the effectiveness of our proposed method over existing state-of-the-art approaches for these problems. The code for the proposed method is available  \href{https://github.com/google-research/google-research/tree/master/implicit_constrained_optimization}{at this url}.
\end{abstract}

%% file: intro.tex
\section{Introduction}
\label{sec:inro}
In many modern machine learning applications, the performance of a model is evaluated using metrics that are complex and nuanced. For example, in retrieval systems, it is common to evaluate a scoring model based the area under the precision-recall curve, or the ROC curve, or on its precision at a certain recall value \cite{eban2017scalable}. Similarly, in many medical diagnostic applications, a model is required to yield low false positive rates while restricting its false negatives rate to be within an allowed limit \cite{rao2008kdd}, while in machine learning fairness applications one might be interested in imposing the ``80\% rule'', which requires a positive prediction rate of at least 80\% on the minority class \cite{biddle2006adverse, zafar2017fairness}.

The above problems cannot be directly solved by minimizing a standard classification loss. In fact, prior work has found that doing so can result in inferior model performance \cite{joachims2005support,koyejo2014consistent,kar2014online,eban2017scalable, cotter2019optimization}. Moreover, many of the metrics that we are interested in have a non-decomposable structure, \ie, they cannot be expressed directly in terms of an average over individual data points, making them hard to optimize using standard optimization tools. Much prior work has sought to address this problem, resulting in a range of methods targeting different classes of non-decomposable metrics \cite{yue2007support, narasimhan2013svmpauctight, kar2015surrogate, yan2018binary}.

In this paper, we consider a popular family of non-decomposable objectives that have a certain thresholded form. This includes metrics like the false negative rate (FNR) at a certain fixed false positive rate (FPR), precision at a fixed recall, precision@$K$, AUC-PR, and AUC-ROC, as well as more recent threshold-based fairness metrics \cite{hardt2016equality}. The task of optimizing these metrics can naturally be written as a constrained optimization problem, wherein one seeks to optimize a quantity such as the model's precision or false positive rate at one or more thresholds, subject to the model satisfying a set of rate constraints at those thresholds. The dominant approach for solving such rate-constrained problems has been to relax the constraints with surrogate losses, and to formulate an equivalent Lagrangian-based primal-dual problem \cite{eban2017scalable}. Follow-up work has improved upon this approach by using the surrogate relaxations only for the primal updates, but not the dual \cite{cotter2019two,narasimhan2019optimizing}.

Our proposed optimization strategy departs significantly from the prior Lagrangian-based methods, and avoids explicitly solving the constrained optimization problem. Instead, we express the threshold variables in the optimization problem as an \textit{implicit} function of the model parameters, and thus re-formulate the problem as an unconstrained optimization over the model parameters. By appealing to the Implicit Function Theorem \cite{tu2011introduction}, we show how to compute the gradients for the resulting unconstrained objective, despite not knowing the form of the implicit function, and then use them to perform standard gradient-based optimization (see Section \ref{sec:method}). Although the Implicit Function Theorem makes a local statement about the existence of the implicit function (\ie, valid in a small neighborhood around current model parameters), we can still effectively use the theorem to make local gradient updates towards optimizing the objective.  

We experiment with two image classification datasets and several UCI datasets, and show  that our proposed method often performs significantly better than the state-of-the-art constrained optimization solvers in optimizing popular metrics such as FNR at fixed FPR, and the ``partial'' areas under the ROC and Precision-Recall curves evaluated at a selection range of FPR/recall values (see Section \ref{sec:exp}). We find our approach to be particularly effective when used to target extreme values of FPR or recall. We also discuss how our formulation can be extended to apply to more complex learning problems, such as query-based ranking, where standard constrained optimization techniques are known to have notable drawbacks (see Section \ref{sec:discussion}).

%% file: formulation.tex
\section{Problem Formulation}
\label{sec:formulation}
We describe our formulation in a binary classification setting, with input space $X$ and binary labels $\{0,1\}$. Later, we will discuss how to extend our setup to   multi-class classification problems. Our goal is to learn a scoring model $s^\theta: X \> \R$, parameterized by $\theta \in \R^p$, whose scores can be thresholded to make a binary prediction. We denote a scoring model thresholded at $t$ by $s^\theta_t(x) = \1(s^\theta(x)> t)\in\{0,1\}$. We will use $\tp(s^\theta_t)$, $\fp(s^\theta_t)$ and $\fn(s^\theta_t)$ to denote the true positives, false positive and false negatives respectively for the thresholded classifier.

We are interested in solving constrained optimization problems of the form:
\begin{equation}
\min_{\theta \in \R^p} f(\theta,\lambda) \quad \text{s.t. }~~ g(\theta,\lambda)=\0
\label{eq:opt}
\end{equation}
where the objective $f: \R^p \times \R^m \> \R$ maps the model parameters 
$\theta \in \R^p$ and a set of $m$ thresholds $\lambda \in \R^m$  to a real value, and  the constraints $g: \R^p \times \R^m \> \R^m$ map $(\theta, \lambda)$ to $m$ real numbers. We further assume $\theta$ stays in the feasible region, \ie, $\forall \theta,\,\, \exists \lambda\,\, \text{ s.t. } g(\theta,\lambda)=\0$. We will further discuss how several commonly used constrained optimization problems of interest in machine learning satisfy this assumption of feasibility. We provide some popular examples of evaluation metrics below.

\begin{exmp}[\textbf{Precision at fixed recall}]
\label{ex:patr}
To maximize the model's precision at the threshold $\lambda \in \R$ at which its recall is $\beta$, we will have: 
\begin{eqnarray*}
f(\theta,\lambda) &=& -\prec(s^{\theta}_\lambda) = -\frac{\tp(s^\theta_\lambda)}{\tp(s^\theta_\lambda)+\fp(s^\theta_\lambda)};\\
g(\theta,\lambda) &=& \rec(s^{\theta}_\lambda) - \beta = \frac{\tp(s^\theta_\lambda)}{\tp(s^\theta_\lambda)+\fn(s^\theta_\lambda)}-\beta.
\end{eqnarray*}
\end{exmp}

\begin{exmp}[\textbf{FNR at fixed FPR}]
To minimize the model's false negative rate at the threshold $\lambda \in \R$ at which its false positive rate is $\beta \in [0,1]$, we will have: 
\begin{eqnarray*}
f(\theta,\lambda) &=& \fnr(s^\theta_\lambda) ~=~
\frac{\fn(s^\theta_\lambda)}{\tn(s^\theta_\lambda)+\fp(s^\theta_\lambda)};\\
g(\theta,\lambda) &=& \fpr(s^\theta_\lambda) ~=~ \frac{\fp(s^\theta_\lambda)}{\tp(s^\theta_\lambda)+\fn(s^\theta_\lambda)}-\beta.
\end{eqnarray*}
\end{exmp}

\begin{exmp}[\textbf{Precision at $k$}]
\label{ex:patr}
To maximize the model's precision at the threshold $\lambda \in \R$ at which it achieves a coverage of $k$, we can set:
\begin{eqnarray*}
f(\theta,\lambda) &=& -\prec(s^{\theta}_\lambda);\\
g(\theta,\lambda) &=& \tp(s^\theta_\lambda)+\fp(s^\theta_\lambda)-k.
\end{eqnarray*}
\end{exmp}

\begin{exmp}[\textbf{AUC-PR}]
To maximize the area under the Precision-Recall curve, following \citet{eban2017scalable}, we use a Riemann approximation to the area: we divide the recall range into $m$ equally-spaced values $\beta_1,\ldots,\beta_m \in [0,1]$, and evaluate the average precision that the model achieves when thresholded to match each of the target recalls. This can be written as a constrained optimization problem with $m$ thresholds $\lambda_1, \ldots, \lambda_m \in \R$, and with the objective and constraints set to:
\begin{eqnarray*}
f(\theta,\lambda) &=&
-\frac{1}{m}\sum_{i=1}^m
\prec(s^\theta_{\lambda_i})\\
 g_i(\theta,\lambda) &=& 
\rec(s^\theta_{\lambda_i})
-\beta_i, 
\forall i \in [m].
\end{eqnarray*}
One can similarly compute the ``partial'' area under the PR curve in any given range of recall (precision) targets by thresholding only at those particular recall (precision) values. This is particularly useful for excluding low recalls (precisions), since one is generally uninterested in the performance of the model at such thresholds.
\end{exmp}

\begin{exmp}[\textbf{AUC-ROC}]
\label{ex:auc-roc}
To maximize the (partial) area under the Receiver Operator Characteristic (ROC) curve, we can again apply a Riemann approximation: we divide the FPR range into $m$ values $\beta_1,\ldots,\beta_m$, and compute the average TPR at $m$ thresholds $\lambda_1,\ldots, \lambda_m \in \R$, chosen to satisfy the FPR targets:
\begin{eqnarray*}
f(\theta,\lambda) &=&
-\frac{1}{m}\sum_{i=1}^m
\tpr(s^\theta_{\lambda_i})\\
 g_i(\theta,\lambda) &=& 
\fpr(s^\theta_{\lambda_i})
-\beta_i, 
\forall i \in [m].
\end{eqnarray*}
\end{exmp}

\begin{exmp}[\textbf{Fairness criterion}]
\label{ex:fairness}
In a typical group fairness application \cite{hardt2016equality}, each example belongs to one of $m$ protected groups, and the goal is to constrain the model to have equitable performance across all groups. One way to enforce this requirement is to introduce a separate threshold $\lambda_i$ for examples from each group, and to tune  them to satisfy the fairness constraints. 
For example, the popular \emph{demographic parity} constraint for two (disjoint) groups, which requires equal positive prediction rates for both groups, can be encoded as:
\begin{eqnarray*}
g(\theta,\lambda) &=&  \frac{1}{m_1}\left({\tp_1(s^\theta_{\lambda_1}) + \fp_1(s^\theta_{\lambda_1})}\right) \\
&&~~~~~-\frac{1}{m_2}\left({\tp_2(s^\theta_{\lambda_2}) + \fp_2(s^\theta_{\lambda_2})}\right),
\end{eqnarray*}
where $\tp_1, \tp_2$ are the true positives on examples belonging group $1$ and $2$ respectively, $\fp_1, \fp_2$ are the false positives for the two groups, and $m_1$ and $m_2$ are the number of examples in the two groups.
\end{exmp}
{\bf Feasibility of constraints by tuning $\lambda$.~} Assuming that the model does not map two different training examples to same outputs, it is easy to see that for fixed model parameters $\theta$, any of the aforementioned rate constraints (\eg, false positive rate, precision, recall, \etc) can be satisfied to any feasible value by tuning \emph{only} the thresholds $\lambda_1, \ldots, \lambda_m$. 

All the rate based optimization objectives and constraints discussed above are non-smooth. To make the problem amenable to gradient based optimization, following earlier work \citep{eban2017scalable,cotter2019two,narasimhan2019optimizing} we replace $f$ and $g$ with smooth differentiable surrogates $\tilde{f}$ and $\tilde{g}$, and relax \eqref{eq:opt} into:
\begin{align}
\min_{\theta \in \R^p} \tilde{f}(\theta,\lambda) \quad \text{s.t. }~~ \tilde{g}(\theta,\lambda)=\0.
\label{eq:opt-relaxed}
\end{align}

{\bf Surrogate losses.~} We use sigmoid and softplus functions, denoted as $\sigma(\cdot)$, as surrogates in our experiments. Specifically, we replace the innermost indicators with smooth surrogate $\sigma$. For example, if the objective $f$ is FNR = ($\frac{\text{FN}}{\text{total  positives}}$) and $p_i$ is the prediction (logit) for $i$th example, then we replace FN $=\sum_{i:y_i=1}\mathbb{I}_{p_i<\lambda}$ with $\widetilde{\text{FN}}=\sum_{i:y_i=1}\sigma_
\tau(-(p_i-\lambda))$, yielding the surrogate $\tilde{f}=\frac{\widetilde{\text{FN}}}{\text{total positives}}$ (where $\sigma_\tau(x)=\sigma(\tau x)$ denotes a temperature scaled sigmoid or softplus function). We use similar surrogates for ratio based objectives such as precision and recall. For example, if $f$ is precision ($\frac{\text{TP}}{\text{predicted   positives}}$) then we replace
TP=$\sum_{i:y_i=1}\mathbb{I}_{p_i>\lambda}$ with
$\widetilde{\text{TP}}=\sum_{i:y_i=1}\sigma_\tau(p_i-\lambda)$ and predicted-positives $=\sum_i \mathbb{I}_{p_i>\lambda}$ with $\widetilde{\text{PP}}=\sum_i\sigma_\tau(p_i-\lambda)$, yielding the surrogate $\tilde{f}=\frac{\widetilde{\text{TP}}}{\widetilde{\text{PP}}}$.

%% file: method.tex
\section{Optimization with Implicit Thresholds}
\label{sec:method}
The canonical approach to solving the constrained optimization problem in \eqref{eq:opt-relaxed} is to formulate a Lagrangian for the problem, and then perform gradient updates to maximize the Lagrangian over the mulitipliers and minimize it over $\theta$ and $\lambda$. Our key idea is to avoid explicitly solving the constrained problem by instead formulating an equivalent \emph{unconstrained} problem in which we express the thresholds $\lambda$ as an \emph{implicit} function of the model parameters $\theta$ (within a neighborhood around $\theta$).

To this end, we make use of the Implicit Function Theorem \citep{tu2011introduction}. 
Specifically, suppose the point $(\theta^0,\lambda^0)\in\reals^{p+m}$ satisfies the constraint, \ie,  $\tg(\theta^0,\lambda^0)=0$. 
Then we can express the thresholds as $\lambda^0=\th(\theta^0)$ in a neighborhood around $\theta^0$, for
some implicit function $\th$:

\begin{theorem}[\textbf{Implicit Function Theorem} \cite{tu2011introduction}]
\label{thm:implicit-function}
Let $U$ be an open subset in $\reals^p\times\reals^m$ and $\tg:U\to \reals^m$ a $C^1$ map. Write $(\theta,\lambda)$ for a point in $U$, with $\theta\in\reals^p, \lambda\in\reals^m$. At a point $(\theta^0,\lambda^0)\in U$ where $\tg(\theta^0,\lambda^0)=0$ and the determinant $\det[\partial \tg^i/\partial\lambda^j(\theta^0,\lambda^0)]$ is nonzero, there exists a neighborhood $\Theta\times\Lambda$ of $(\theta^0,\lambda^0)$ in $U$ and a unique $C^1$ function $\th:\Theta\to\Lambda$ such that in $\Theta\times\Lambda\subset U\subset\reals^p\times\reals^m$, 
\[
\tg(\theta,\lambda)=0\quad \Longleftrightarrow\quad \lambda=\th(\theta)
\]
\end{theorem}

Using this theorem, we can write the implicit threshold as $\lambda=\th(\theta)$ 
within a
neighborhood around $\theta^0$,
which enables us to turn problem  \eqref{eq:opt-relaxed} into the equivalent 
unconstrained problem of minimizing $\tf(\theta,\th(\theta))$. %

\subsection{Characterization of the implicit function}
A differentiable function 
$\th(\theta) \in \R^m$ that
provides us with the $m$ thresholds at which $\tilde{f}(\theta,\th(\theta))=\0$
may not always exist,
and even if it does, it may be  available in closed-form only in some highly simplified settings.\footnote{For example, if the distribution of the instances $x \in \mathbb{R}^p$ conditioned on label $y = 1$ is a Gaussian distribution with mean $\mu \in \mathbb{R}^p$ and covariance matrix $\Sigma \in \mathbb{R}^{p\times p}$, and suppose we wish to constrain the false negative rate (FNR) for a linear model parameterized by $\theta \in \mathbb{R}^p$. Then the FNR at any  threshold $\lambda \in \R$ is given by $\P_{x \sim \mathcal{D}_+}\left(\theta^\top x \leq \lambda\right)$, and the threshold $\lambda$ at which the FNR is $\beta$ is given by $\th(\theta) = \Phi^{-1}(\beta)$, where $\Phi$ is the CDF of a normal distribution with mean $\theta^\top \mu$ and variance $\theta^\top \Sigma \theta$.}
Nonetheless, under some assumptions,
we can show that when $\th$ exists, the resulting composite function $\tilde{f}(\theta,\tilde{h}(\theta))$ is convex in $\theta$.
\begin{prop}
\label{prop:convexity}
Let $m=1$. Suppose the objective $\tilde{f}(\theta, \lambda)$ 
is jointly convex in $(\theta, \lambda)$ and is strictly increasing in $\lambda \in \R$, 
and the constraint 
$\tg(\theta, \lambda)$ is jointly convex in $(\theta, \lambda)$ and is strictly descreasing in $\lambda \in \R$.
Suppose there exists a $C^1$ function $\th: \R^p \> \R$ such that $\tg(\theta, \th(\theta)) = 0, \forall \theta$. Then $\th$ is convex in $\theta$. Consequently, the composite objective $\tilde{f}(\theta, \tilde{h}(\theta))$ is convex in $\theta$.
\end{prop}

The proof adapts a result from \citet{wurker2001convexity}
and is given in
Appendix A. %
The assumptions in the proposition hold
for simple linear models, for example, when  minimizing the FPR while constraining the FNR (see appendix for details).

\subsection{Gradient computation}
To compute a (local) derivative for $\tf(\theta,\th(\theta))$ w.r.t. $\theta$ within the neighborhood of $\theta^0$ in Theorem \ref{thm:implicit-function},
we use:
\begin{align}
\nabla_\theta \tf(\theta,\th(\theta)) = \nabla_\theta \tf(\theta,\lambda) + \frac{\partial \tf(\theta,\lambda)}{\partial \lambda} \nabla_\theta \th(\theta),
\label{eq:diffobj}
\end{align}
where for simplicity we show the derivative for a scalar $\lambda$. 
We will further need the derivative of the implicit function $\th(\cdot)$. Since $\tg(\theta,\th(\theta))=0$ in this neighborhood, we have 
\begin{align}
\begin{split}
& \nabla_\theta \tg(\theta,\lambda) + \frac{\partial \tg(\theta,\lambda)}{\partial \lambda} \nabla_\theta \th(\theta) = 0 \\
\Longrightarrow & \nabla_\theta \th(\theta) = - \frac{\nabla_\theta \tg(\theta,\lambda)}{\frac{\partial \tg(\theta,\lambda)}{\partial \lambda}}
\end{split}
\label{eq:diffcons}
\end{align}
This gives us the derivative of the implicit function which can be plugged into Eq. \eqref{eq:diffobj} to get the final gradients for model parameters $\theta$.

\begin{figure}[t]
\begin{algorithm}[H]
\caption{Implicit Constrained Optimization (ICO)%
}
\label{algo:implicit}
\begin{algorithmic}[1]
\STATE \textbf{Hyper-parameters:} %
OPT, $\tau \in \mathbb{N}$
\STATE \textbf{Intialize:} $\theta^0, \lambda^0$
\STATE \textbf{For} $t = 1 ~\text{to}~ T$:
\STATE ~~~~$H^t = - {\nabla_\theta \tg(\theta^t,\lambda^t)}/{\frac{\partial \tg(\theta^t,\lambda^t)}{\partial \lambda}}$
\STATE ~~~~$G^t = \nabla_\theta \tf(\theta^t,\lambda^t) + \frac{\partial \tf(\theta^t,\lambda^t)}{\partial \lambda}  H^t$
\STATE ~~~~$\theta^{t+1} = \text{OPT}(\theta^t, G^t)$ ~~~~~\text{//\,optimizer step}
\STATE ~~~~\textbf{If} $t ~\text{mod}~ N = 0$: ~~~~~~~~~~~~~\text{//\,correction step}
\STATE ~~~~~~~~\text{Set $\lambda^{t+1}$ s.t. $g(\theta^{t+1},\lambda^{t+1}) = \0$}
\STATE ~~~~\textbf{Else}:~~~~~~~~~~~~~~~~~~~~~~~ \text{//\,gradient based update for $\lambda$}
\STATE ~~~~~~~~$\lambda^{t+1} = \lambda^{t} + \langle \nabla_\theta \th(\theta^t), \theta^{t+1} - \theta^t\rangle $
\STATE ~~~~\textbf{End If}
\STATE \textbf{End For}
\STATE \textbf{Return} $\theta$
\end{algorithmic}
\end{algorithm}
\end{figure}

\subsection{Updating thresholds}
Having performed the gradient update on $\theta$ with:
$$\theta^{t+1} = \theta^{t} - \eta\nabla_\theta\tf(\theta^t,\th(\theta^t)),$$ 
where $\eta > 0$ is a step-size parameter, what remains is to update the threshold. %
Again appealing to the Implicit Function Theorem, in the neighborhood around the current iterate $\theta^{t}$, we can approximate the new threshold as: \begin{eqnarray*}
\lambda^{t+1} &=& \th(\theta^{t+1}) = \th(\theta^{t}+\Delta \theta) \\
&\approx & \th(\theta^{t}) + \langle \nabla_\theta \th(\theta^t),\Delta\theta\rangle \\
&=& \lambda^{t} + \langle \nabla_\theta \th(\theta^t),\Delta\theta\rangle.
\end{eqnarray*}
As this is an approximation, we employ a \textit{correction step} after every $N$ minibatch iterations that sets the threshold to satisfy the constraint exactly based on $k$ accumulated minibatches.  
Note that for all the metrics described in Examples \ref{ex:patr}--\ref{ex:auc-roc}, this correction step can be performed efficiently using a straight-forward line search.

\subsection{Practical improvements}\label{sec:method:practical}
Algorithm \ref{algo:implicit} outlines our overall approach. In our experiments, we found it effective to use the unrelaxed (and non-smooth) rates, instead of surrogates, for the correction step, i.e. we set $\lambda^{t+1} = h(\theta^{t+1}),$ where  $h$ computes the threshold at which the unrelaxed $g(\theta^{t+1},\lambda^{t+1})=\0$.

{\bf Regularization.~} In some of our experiments, particularly with smaller UCI datasets, we also impose a \emph{regularizer} on the model parameters that penalizes $\lVert d\tilde{g}(\theta,\lambda) / d\lambda \rVert^2$ w.r.t. $\theta$. This encourages the optimization to prefer model parameters for which the constraint function varies smoothly as a function of the threshold. We expect that this will help the model to generalize better on unseen examples. 

\noindent {\bf Optimizing objective with multiple constraints.~} 
For optimizing objectives involving multiple constraints (and hence multiple thresholds $\lambda=[\lambda_1,\ldots,\lambda_m] \in \R^m$), such as (partial) PR-AUC or ROC-AUC, a na\"ive implementation would need multiple gradient computations w.r.t. $\theta$ as follows:
\begin{align}
\begin{split}
\nabla_\theta\tf(\theta,h_1(\theta),\ldots& ,h_m(\theta_m)) 
= \nabla_\theta  \tf(\theta,\lambda)  - \\ & \sum_i \frac{\partial \tf(\theta,\lambda)}{\partial \lambda_i}  \frac{\nabla_\theta \tg(\theta,\lambda_i)}{\frac{\partial \tg(\theta,\lambda_i)}{\partial \lambda_i}}
\end{split}
\end{align}
It needs computation of $\nabla_\theta \tg(\theta,\lambda_i)$ for all $m$ constraints. We avoid this by first computing the partial derivatives of $\tf(\theta,\lambda)$ and $\tg(\theta,\lambda_i)$ w.r.t. $\lambda_i$ which are much cheaper to compute, and then treating their ratios as constants $r_i$ (akin to using \emph{stop-gradient}). Denoting $h(\theta)=\{h_1(\theta),\ldots,h_m(\theta_m)\}$, we then rewrite the gradient computation as 
\begin{align}
\begin{split}
\nabla_\theta\tf(\theta,h(\theta)) = \nabla_\theta\tf(\theta,\lambda) -  \nabla_\theta \sum_i  r_i \tg(\theta,\lambda_i),
\end{split}
\end{align}
\noindent which again reduces to just two gradient computations. We also disable the gradient based updates for thresholds to avoid computing separate $\nabla_\theta \tg(\theta,\lambda_i)$ for all $i$, and only rely on the correction step of Algorithm \ref{algo:implicit} every $\tau$ minibatches.

%% file: related.tex
\section{Related Work}
\label{sec:related}

The problem of training models to optimize a given non-decomposable metrics has received much attention in the literature. Early methods on this topic focused on constructing convex surrogates that closely approximate the metric of interest
\cite{joachims2005support, yue2007support, narasimhan2013structural, kar2014online,  mohapatra2014efficient, narasimhan2015optimizing}, often using 
structured support vector machines \cite{tsochantaridis2005large}. One of the drawbacks of these approaches is that they are not directly amenable to handling constraints on multiple rates, and may sometimes result in a loose approximation to the metric \cite{kar2015surrogate}. More recent methods seek to directly optimize a given rate metric subject to constraints on multiple rate metrics
\cite{goh2016satisfying, eban2017scalable, narasimhan2018learning, cotter2019optimization, cotter2019two, narasimhan2019optimizing}, and come with scalable  gradient-based solvers. %
There has also been work on construction of structured surrogates \cite{fathony2020ap,bao2020calibrated}. 

There is also a distinction between methods which handle classification metrics such as the F-measure \cite{koyejo2014consistent, narasimhan2014statistical, yan2018binary}, where often tuning a threshold on a pre-trained class probability model results in a consistent estimator, and those that handle scoring metrics such as the precision-recall and AUC metrics we consider in this paper \cite{eban2017scalable}, where the focus is on learning a scoring model that performs well at one or more operating thresholds. Other techniques focus on optimizing specialized evaluation metrics that, for example, emphasize good top-$k$ performance in ranking and classification tasks \cite{agarwal2011infinite, Boyd+12,fan2017learning,lapin2017analysis,hiranandani2020optimization}.

Our approach is most closely related to the method of \citet{eban2017scalable}, who encode the given metric as  constraints on classification rates, relax the rates with differentiable surrogates, and perform gradient updates to minimize over the model parameters, and maximize over the Lagrangian multipliers for the constraints. The recent work of \citet{cotter2019two} and \citet{narasimhan2019optimizing} improves upon their method by observing that the use of surrogates is \emph{only} required when minimizing over the model parameters, while the maximization over the Lagrange multipliers can be performed with the original unrelaxed rates. The resulting min-max problem can be interpreted as a non-zero-sum game, for which the authors provide efficient gradient-based algorithms to find an equilibrium. This selective use of surrogates is incorporated into our proposal: we use surrogates only when we need to compute gradients for the objective $\tf(\theta, \th(\theta))$, whereas, as we mentioned in Section \ref{sec:method:practical}, we use the original unrelaxed rates while computing a correction for the threshold $\lambda = h(\theta)$.

Unlike these earlier papers, our method avoids explicitly solving a constrained optimization problem, and instead expresses the threshold as an implicit function of the model parameters. %
This proposal is similar in flavor to the approach taken by \citet{mackey2018constrained}, who like us formulate an unconstrained objective, but do so by expressing the threshold as a quantile of the model scores. 

Finally, the growing literature on fairness in machine learning has opened the door for many new applications for constrained optimization \cite{hardt2016equality, agarwal2018reductions}, introducing many group-based fairness metrics that can be easily handled using the proposed approach. %

%% file: experiments.tex
\section{Experiments}
\label{sec:exp}

\begin{table}[]
    \centering
        \caption{Summary of datasets.} 
    \begin{tabular}{cccc}
	\toprule[0.25ex]
         \textbf{Datasets} & \textbf{\#Examples}& \textbf{\#Features}\\
    \midrule[0.25ex]
    CelebA & 202,599 & 32$\times$32$\times$3 \\
    BigEarthNet &590,326 & 40$\times$40$\times$ 3\\
        \hline
    Letter& 19,999 & 16\\ %
    IJCNN1 &  49990 & 22\\ %
    Adult & 48,842 & 122\\ %
    Spambase & 4,601 & 57\\ %
    Com.\ \& Crime & 1,994 & 145\\ %
	\toprule[0.25ex]
    \end{tabular}
    \label{tab:data}
\end{table}

\begin{table*}[t]
	\centering
	\caption{Minimizing false negative rate (FNR) at a given false positive rate (FPR) for {\bf CelebA}.~~ The mean FNR (in \%) are reported over five random trials for cross-entropy/\,TFCO/\,\proposed, respectively. Proposed \proposed~outperforms both CE and TFCO by a considerable margin. We report results on more attributes along with the std. errors in Appendix  \ref{app:celeba}. 
	\textit{Lower} values are better.}
	\vspace{2mm}
	\begin{tabular}{c c c c c c c}
	\toprule[0.25ex]
	\parbox[t]{2mm}{\rotatebox[origin=c]{90}{FPR}} & High-cheekbones & Heavy-makeup & Wearing-lipstick & Smiling & Black-hair & Blond-hair \\[1mm]
	 \midrule
	1\% & 53.5/\,49.0/\,{\bf 46.9}& 57.0/\,57.0/\,{\bf 49.6}& 44.0/\,42.6/\,{\bf 37.5}& 37.4/\,35.9/\,{\bf 33.7}& 69.3/\,64.4/\,{\bf 63.2}& 40.4/\,38.6/\,{\bf 36.8}  \\
	
	2\% & 44.8/\,40.9/\,{\bf 39.8}& 45.6/\,41.2/\,{\bf 38.9}& 32.7/\,30.4/\,{\bf 26.7}& 29.4/\,27.8/\,{\bf 26.1}& 56.4/\,52.0/\,{\bf 50.5}&  28.9/\,25.6/\,{\bf 24.2} \\ 
	
	5\% & 32.9/\,30.1/\,{\bf 28.5}& 28.2/\,25.4/\,{\bf 23.1}& 16.3/\,14.9/\,{\bf 13.1}& 18.7/\,17.0/\,{\bf 16.9}& 36.7/\,32.4/\,{\bf 32.5}& 13.4/\,11.6/\,{\bf 10.8} \\
	
	10\% & 22.9/\,20.4/\,{\bf 19.7}& 15.1/\,13.6/\,{\bf 12.4}& 6.6/\,5.9/\,{\bf 4.7} & 11.7/\,10.7/\,{\bf 10.2} & 23.0/\,19.2/\,{\bf 18.6}& 6.5/\,4.9/\,{\bf 4.7} \\
	\bottomrule[0.25ex]
	\end{tabular}
	\label{tab:celeba_frr}
\end{table*}

\begin{table*}[t]
	\centering
	\caption{Maximizing area under the ROC curve for {\bf CelebA}, in a given FPR range $[0,\beta]$ for $\beta\in\{1\%,2\%,5\%,10\%,20\%\}$.~~ The mean ROC-AUC are reported over five random trials for cross-entropy/\,TFCO/\,\proposed, respectively. Last column shows the mean partial AUC over all 8 attributes.  We report results on more attributes along with the std. errors in Appendix \ref{app:celeba}.
	\textit{Higher} values are better.}
	\vspace{2mm}
	\begin{tabular}{c c c c c c |c}
	\toprule[0.25ex]
	{$\bm{\beta}$} & High-cheekbones & Heavy-makeup & Wearing-lipstick & Smiling & Black-hair & Mean \\[1mm]
	 \midrule
	1\% & 66.1/\,62.9/\,{\bf 69.8}& 65.0/\,{\bf 68.5}/\,66.7&70.2/\,{\bf 74.8}/\,72.3& 75.4/\,{\bf 78.0}/\,75.6& 60.5/\,{\bf 61.4}/\,61.2&  64.7/\,65.9/\,{\bf 66.1}\\
	
	2\% & 70.8/\,{\bf 74.9}/\,73.2 & 68.8/\,{\bf 73.4}/\,71.6 & 75.4/\,{\bf 79.3}/\,78.4& 78.4/\,{\bf 81.5/}\,79.8 & 64.5/\,{\bf 66.5}/\,66.1&  68.6/\,{\bf 70.8}/\,70.5\\ 
	
	5\% &75.9/\,73.8/\,{\bf 78.5} &75.5/\,{\bf 79.5}/\,78.3& 82.2/\,84.7/\,84.6&  83.5/\,65.0/\,{\bf 84.8}& 70.9/\,73.2/\,{\bf 73.8} & 74.7/\,72.8/\,{\bf 76.8}\\
	
	10\% & 80.1/\,74.1/\,{\bf 82.7}& 81.5/\,{\bf 85.0}/\,84.4 & 87.7/\,89.3/\,{\bf 89.7 }&  86.7/\,73.8/\,{\bf 88.8}& 78.0/\,79.9/\,{\bf 80.2}&  79.7/\,77.9/\,{\bf 81.8}\\
	
	20\% & 84.2/\,72.4/\,{\bf 86.8}& 88.2/\,89.9/\,89.8& 91.8/\,93.1/\,{\bf 93.9}& 90.9/76.6/{\bf 92.1}& 84.3/86.0/86.1& 84.9/\,82.4/\,{\bf 86.8}\\
	\bottomrule[0.25ex]
	\end{tabular}
	\label{tab:celeba_auc}
\end{table*}

We evaluate our approach on five UCI  classifications tasks, and two %
image classification tasks. A summary of the datasets used in the main text is provided in Table \ref{tab:data}. We present additional experimental results in Appendices %
B, C and D. 

\textbf{Baselines.}~ We compare the  proposed \proposed~approach with the state-of-the-art tools of \citet{cotter2019optimization} and \citet{narasimhan2019optimizing} for constrained optimization, open-sourced as a part of the TensorFlow Constrained Optimization Library (TFCO)\footnote{\url{https://github.com/google-research/tensorflow_constrained_optimization}}. The  prior technique of \citet{eban2017scalable} can be viewed as a special case of the functionality provided by this library. We also compare to the baseline approach of optimizing a standard cross-entropy (CE) loss.

\textbf{Experimental protocol.}~ For our experiments with Image datasets, we use a 6-layer neural network with 5 convolutional layers with 128, 256, 256, 512 and 512 filters respectively. We use ReLU activation functions and batch normalization layers in the network. 
We use a separate validation split for model selection in all our experiments. For all three methods (CE, TFCO, and \proposed), we use the evaluation metric of interest on the validation set for model selection (\eg, FNR, ROC-AUC, PR-AUC). This makes the CE baseline further strong in our experiment.
We do 5 random trials for each experiment and report the average value of the metric. Other details such as standard deviations for the random trials are reported in the Appendix. 

\subsection{Minimizing FNR at FPR $\beta$}
First, we consider the task of minimizing false negative rate (FNR) at a given false positive rate (FPR) of $\beta$. This setting is particularly relevant for security sensitive applications where one wants to operate at a desired false positive rate. We experiment with the publicly available  CelebA dataset \cite{liu2015deep}, which contains 202,599 celebrity face images of size $32\times 32 \times 3$. CelebA has 40 annotated binary attributes for every face image, of which we randomly choose $8$  for our experiments. We use the standard train, validation, and test splits\footnote{\url{https://www.tensorflow.org/datasets}} for CelebA, and train a binary classifier for each attribute. 

We use TFCO and \proposed~for optimizing FNR at four different FPR targets: 1\%, 2\%, 5\% and 10\%.
For the cross-entropy baseline, we use  Adam optimizer~\citep{Kingma:2015} with a learning rate of 0.001. For TFCO, we use Adam for both the primal and dual updates; the primal learning rate is set to 0.001, while the dual learning rate is chosen from $\{0.1, 0.01\}$ using the validation sample. We use the cross-entropy surrogate  (softplus function for binary case) provided by the TFCO library to approximate the rates. 
 For \proposed, we again use  Adam  with a learning rate of 0.001. We approximate the rates for \proposed~with temperature-scaled sigmoid surrogates, and choose the temperature parameter from the range $\{0.001, 0.01, 0.1, 1.0\}$ using the validation sample. We do not apply gradient regularization, and 
  perform the correction step described in Section \ref{sec:method:practical} once every $1000$ mini-batch updates using data from next $10$ mini-batches.
All optimizers use a batch size of 512.

\begin{table}[t]
	\centering
	\caption{Maximizing area under the ROC curve for {\bf BigEarthNet}, in a given FPR range $[0,\beta]$ for $\beta\in\{5\%,10\%,20\%\}$.~~ The mean ROC-AUC are reported over five random trials for cross-entropy/\,TFCO/\,\proposed, respectively. We report results on more attributes along with the std. errors in Appendix \ref{app:bigearth}. 
	\textit{Higher} values are better.}
	\vspace{2mm}
 	\footnotesize
	\begin{tabular}{c c c c}
	\toprule[0.25ex]
	\multirow{3}{*}{Labels} & \multicolumn{3}{c}{$\bm{\beta}$}\\
	\cmidrule{2-4}
	 & 5\% & 10\% & 20\%  \\[1mm]
	 \midrule
	BLF & 66.2/\,66.4/\,{\bf 69.9}& 71.0/\,71.7/\,{\bf 73.9}& 75.4/\,76.2/\,{\bf 77.9}  \\
	CC & 62.2/\,62.7/\,{\bf 63.6} & 67.4/\,66.8/\,{\bf 68.3} & 73.8/\,{\bf 76.0}/\,74.9 \\
	CF & 71.9/\,71.5/\,{\bf 74.7}& 78.6/\,80.0/\,{\bf 80.7}& 84.8/\,{\bf 86.6}/\,86.0  \\
	DUF & 69.8/\,71.8/\,{\bf 73.9} & 75.1/\,77.2/\,{\bf 78.0} & 78.8/\,81.4/\,{\bf 81.8} \\ 
	ANV & 58.8/\,58.8/\,{\bf 60.4}& 62.7/\,63.9/\,{\bf 64.4}& 67.8/\,69.1/\,69.0 \\
	\midrule
	Mean & 65.8/\,66.2/\,{\bf 68.5} & 71.0/\,71.9/\,{\bf 73.1}& 76.1/\,77.9/\,77.9 \\
	\bottomrule[0.25ex]
	\end{tabular}
	\label{tab:bigearth_auc}
\end{table}

We present in Table \ref{tab:celeba_frr} the test evaluation metrics. The proposed \proposed~method performs the best for all six prediction tasks and for all four FPR targets, with TFCO coming in second.
Interestingly, the gap between \proposed~and the other methods is the larger for smaller FPR targets. This suggests that \proposed~is most advantageous when applied to metrics that are harder to optimize. 
Unsurprisingly, cross-entropy optimization often yields higher FNR values at the specified FPR targets than the other methods, indicating that methods which directly optimize performance at the desired FPR do end up performing better at that target. Results on other attributes are reported in Appendix C. %
We also report timing comparisons of ICO and TFCO in Appendix B. %

\begin{figure*}[t]
\centering
\begin{subfigure}{0.49\textwidth}
\centering
\includegraphics[width=0.49\textwidth]{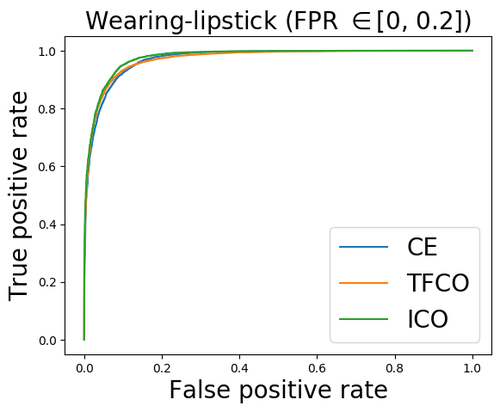}
\includegraphics[width=0.49\textwidth]{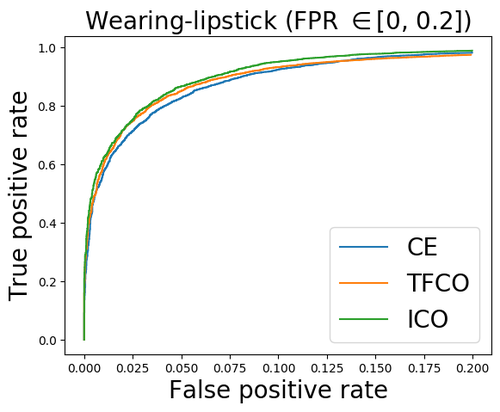}
\caption{Wearing-lipstick attribute}
\end{subfigure}
\begin{subfigure}{0.49\textwidth}
\centering
\includegraphics[width=0.49\textwidth]{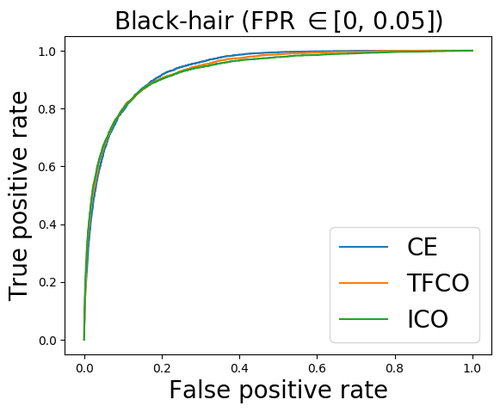}
\includegraphics[width=0.49\textwidth]{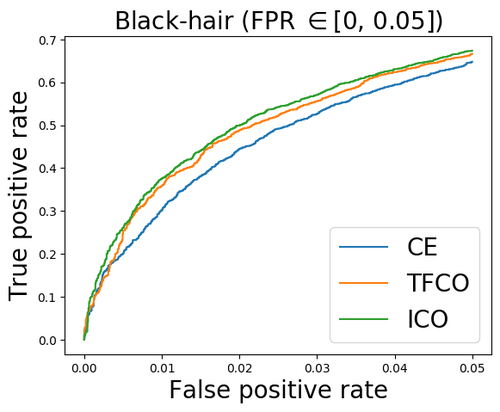}
\caption{Black-hair attribute}
\end{subfigure}

\caption{\emph{ROC curves for CelebA}: {\bf (a)} For attribute \emph{Wearing-lipstick} and optimizing for partial area under the ROC curve with FPR $\in [0,0.2]$, {\bf (b):} For attribute \emph{Black-hair} and optimizing for partial area under the ROC curve with FPR $\in [0,0.05]$. Left figures show full ROC curves while the right figures show the (zoomed-in) ROC curves in the respective target FPR ranges.}
\label{fig:celeba_roc_plots}
\end{figure*}

\begin{figure*}
    \centering
    \begin{subfigure}{0.49\textwidth}
    \centering
    \includegraphics[width=0.99\columnwidth]{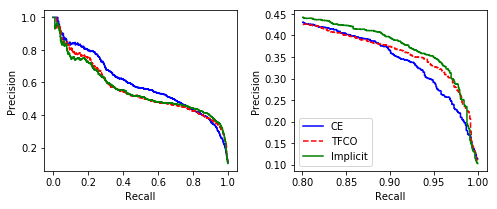}
    \caption{Letter}
    \end{subfigure}
    \begin{subfigure}{0.49\textwidth}
    \centering
    \includegraphics[width=0.99\columnwidth]{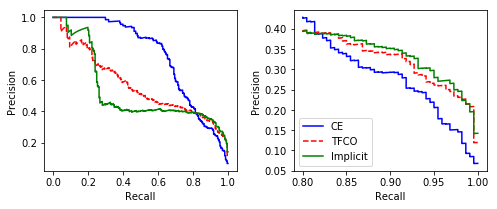}
    \caption{IJCNN1}
    \end{subfigure}
    \caption{Precision-Recall curves on the test set. Both TFCO and the proposed method seek to optimize PR-AUC in the recall range $[0.95,1]$. The left plots for each dataset show the entire curve, while the right plots zoom in on the right-end of the curve.}
    \label{fig:pr-curve}
\end{figure*}

\begin{table*}[]
    \centering
        \caption{Maximizing (partial) PR-AUC in the recall range [0.95, 1] on \textbf{UCI} datasets. Proposed \proposed~performs  better than the other methods on datasets with severe class imbalance. \textit{Higher} values are better.} 
        \vspace{3pt}
    \begin{tabular}{ccccc}
    \toprule[0.25ex]
         & \textbf{\%Positives} &  \textbf{Cross-entropy} & \textbf{TFCO} & \textbf{\proposed}\\
         \hline
    Letter& 4\% & 15.13 $\pm$ 0.86& 20.49 $\pm$ 0.43& \textbf{23.04 $\pm$ 0.77}\\ 
    IJCNN1 & 9.7\% & 21.18 $\pm$ 0.33& 26.14 $\pm$ 0.57& \textbf{27.28 $\pm$ 0.58}\\ 
    Adult & 24\% & 39.74 $\pm$ 0.29& 40.21 $\pm$ 0.38& \textbf{40.34 $\pm$ 0.41}\\ 
    Spam & 39\%& 71.51 $\pm$ 1.73& 73.08 $\pm$ 1.70& \textbf{73.48 $\pm$ 1.80}\\ 
    Com.\ \& Crime & 30\% &
    47.00 $\pm$ 0.94& 47.04 $\pm$ 0.97& 47.03 $\pm$ 1.08
    \\ 
    \bottomrule[0.25ex]
    \end{tabular}
    \label{tab:uci}
\end{table*}

\subsection{Maximizing ROC-AUC in FPR range $[0,\beta]$}
Next, we consider the task of maximizing the (partial) area under the ROC curve in a select range $[0, \beta]$ of FPRs. This metric is used in medical diagnostic tasks and biometric screening \cite{rao2008kdd, ricamato2011partial}, where optimizing performance in a relevant FPR range may prove critical. 
We also compare with a pairwise loss baseline \cite{narasimhan2013svmpauctight}
which optimizes the objective $\frac{1}{N^+|S^-|}\sum_{i:y_i=1} \sum_{j\in S^{-}} \tilde{f}(s^\theta(x_i)-s^\theta(x_j))$, where $s^\theta(x)$ denotes the score (\eg, logits) for example $x$, $N^+$ is the number of positive examples in the minibatch, $S^-$ is the subset of negative examples whose scores lie in the top $\beta$ fraction of all negative examples, and $\tilde{f}$ is the surrogate used for 0-1 loss (either softplus or sigmoid with a temperature hyperparameter as we used for the proposed method). 
We use the pairwise-loss, TFCO, and the proposed method to optimize this metric for five different value of $\beta$: 1\%, 2\%, 5\%, 10\% and 20\%. The results for the pairwise loss baseline are reported in the supplementary material. 
We experiment with CelebA and BigEarthNet \cite{sumbul2019bigearthnet} image datasets. BigEarthNet contains  590,326 Sentinel-2 image patches of size 120$\times$120$\times$3 in RGB, which we down-size to 40$\times$40$\times$3, and 43 annotated binary labels, of which we   choose $5$ for our experiments. These lables are Broad-Leaved Forest (BLF), Complex Cultivation patterns (CC), Coniferous Forest (CF), Discontinuous Urban Fabric (DUF) and Agricultural with Natural Vegetation land (ANV). We split the dataset randomly into 70\% for training, 15\% for validation and 15\% testing.

For both datasets, we train the same convolutional neural network model as the previous experiment. 
Both TFCO and our method divide the specified FPR range $[0,\beta]$ into 10 equally-spaced values, and optimize the average true positive rate (TPR) at those targets. We replicate the same parameter configurations used in the previous experiment, except that the update frequency for \proposed~is performed either once in every 100 updates or 1000 updates, based on which of the two choices yields the highest validation ROC-AUC. We do not use gradient regularization in this experiment. 

We present in Table \ref{tab:celeba_auc}
the test evaluation metrics for different methods on CelebA, where we applied TFCO and \proposed~to optimize the partial ROC-AUC metric for five different values of $\beta$: 1\%, 2\%, 5\%, 10\% and 20\%.
We apply the standard McClish correction \cite{mcclish1989analyzing} to rescale the area between 0 and 100. On at least half the classification tasks, the proposed \proposed~method performs the best, with TFCO coming in second.  On average across all six image attributes, \proposed~is considerably better than TFCO on three of the five values of FPR targets $\beta$. We also show the ROC plots for a few specific cases in Figure \ref{fig:celeba_roc_plots}. 
We present the results for BigEarthNet in Table \ref{tab:bigearth_auc}, where we experiment with three values of $\beta$: 5\%, 10\%, 20\%. For all five labels, the proposed \proposed~is seen to perform better than the baselines for the smaller false-positive ranges, i.e. for smaller $\beta$, thus demonstrating its effectiveness in optimizing performance in the initial portion of the ROC curve for this dataset.
We also report timing comparisons of ICO and TFCO in Appendix B, 
observing that ICO can converge faster than TFCO in terms of wall-clock time.

\subsection{Maximizing PR-AUC in Recall Range $[\beta, 1]$}
In our final set of experiments, we consider the task of maximizing the (partial) area under the Precision-Recall curve in a select range of recall values $[\beta, 1]$. This metric is relevant in retrieval applications, where the quality of the system is often evaluated at multiple recall targets. 
We experiment with the  five smaller datasets in Table \ref{tab:data} obtained from the UCI repository \cite{uci}. %
For the Letter dataset, we treat the most frequent letter as the positive example, and the rest as negative. For the Communities \& Crime dataset, we seek to predict if a community in the US has a crime rate above the $70$th percentile \cite{Kearns+18}. %
We trained a linear model in each case.
We split the datasets into train, validation and test datasets in the ratios 50\%:25\%:25\%. %

Both TFCO and \proposed~divide the specified recall range $[\beta,1]$ into five equally-spaced values, and optimize the average precision at those targets.
We use Adam for the cross-entropy baseline and for TFCO, and Adagrad for the proposed \proposed.
For cross-entropy optimization, we tune the learning rate from the range $\{10^{-3}, 10^{-2}, 10^{-1}, 1\}$, picking the one with maximum PR-AUC metric on the validation sample. For TFCO, we tune the learning rate and dual scale parameters  from the range $\{0.01, 0.1, 1.0\}$ and $\{0.1, 1.0, 10.0\}$  respectively. %
We approximated the  rates for TFCO using the cross-entropy  surrogate loss provided by the library.
For the proposed \proposed, we use a fixed learning rate of 0.1, and approximate the rates with a temperature-scaled sigmoid surrogates, with the temperature parameter for the surrogate chosen from $\{0.5, 1.0, 5.0\}$. We also apply the gradient regularizer described in Section \ref{sec:method:practical}, with the regularization
strength parameter chosen from the range $\{0, 0.05, 0.1\}$. We perform the correction step once every 10 updates.
All optimizers perform full gradient updates.

We first evaluate the performance of different methods at very high recall values. For this, we run both  TFCO and \proposed~to optimize average precision in the recall range $[0.95, 1]$. Table \ref{tab:uci} presents the test PR-AUC metric values in the range $[0.95, 1]$ for the different methods. We find that on the Letter and IJCNN1 datasets, which have severe class imbalance, the proposed approach performs significantly better than the two baselines. On these datasets, cross-entropy optimization performs poorly. On the datasets where the classes are reasonably balanced, all three baselines perform similarly. On the Communities \& Crime dataset, we find our approach yielding better metric values than the other methods on the training sample, but because of the small data size, does not generalize as well to the test set.

\begin{figure}[t]
    \centering
    \begin{subfigure}{0.48\textwidth}
    \centering
    \includegraphics[width=0.7\columnwidth]{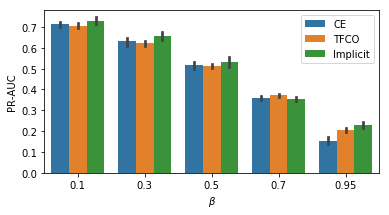}
    \caption{Letter}
    \end{subfigure}
    \begin{subfigure}{0.48\textwidth}
    \centering
    \includegraphics[width=0.7\columnwidth]{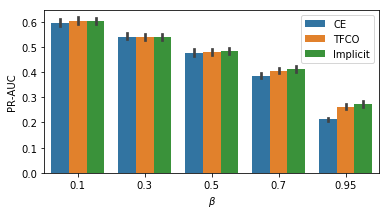}
    \caption{IJCNN1}
    \end{subfigure}
    \caption{Plot of PR-AUC in $[\beta, 1]$ as a function of $\beta$ on the Letter and IJCNN1 datasets. The proposed method is often advantageous for very high $\beta$ values. \textit{Higher} values are better. } %
    \label{fig:prauc}
    
\end{figure}

In Figure \ref{fig:pr-curve}, we show the Precision-Recall cuves for the different methods for the the Letter and IJCNN1 datasets. Notice that while cross-entropy optimization yields higher precision for lower recall values, it does not fair well in the  recall range that matters $[0.95, 1]$. 
Clearly, there is notable benefit to directly optimizing for the recall range that we care about instead of using an off-the-shelf loss function. Moreover, the proposed \proposed~method outperforms TFCO at high recall  values. We also apply the methods to optimize PR-AUC in  recall ranges $[\beta, 1]$, for varying values of $\beta$. In the results shown in Figure \ref{fig:prauc} %
 for the Letter and IJCNN1 datasets, one can see that  the benefit offered by \proposed~is the most benefit over the two baselines is most notable for higher $\beta$ values.

%% file: discussion.tex
\section{Discussion and Future Work}
\label{sec:discussion}
We proposed an approach for optimizing popular constrained optimization problems arising in machine learning that involve non-decomposable rate metrics, such as false positive rate, true positive rate, areas under the precision-recall or ROC curves, \etc. Our approach deviates significantly from the existing methods based on Lagrange multipliers, and uses Implicit Function Theorem to express the classifier thresholds as a function of model parameters. Our experiments showed considerable improvements on optimizing several common evaluation metrics (such as  FNR at a fixed FPR, areas under (partial) precision-recall and ROC curves) over existing state-of-the-art methods \cite{cotter2019optimization,narasimhan2019optimizing} that are part of open-source tools. In this work, we primarily focused on the straightforward setting where the classifier thresholds are expressed as a function of \emph{all} model parameters but it is also possible to consider an alternative setting where the thresholds are expressed as a function of only \emph{a subset} of model parameters (\eg, last few layers of the neural network). We leave this as a direction for future work. 

We close by highlighting how our proposal can be extended to handle more complex settings and constraints.

\subsection{Inequality constraints} 
All the constrained metrics we described are defined as equality constraints. In our current implementation, we handle inequality constraints by searching for thresholds that satisfy the original non-smooth inequality constraints (during the correction step in Algorithm \ref{algo:implicit} and finally at the end of training). However, we can easily extend our proposal to  handle inequality constraints of the form $\tg(\theta,\lambda)\leq \0$ in a more principled manner.
In this case, one can introduce $m$ auxiliary slack variables $\xi_1,\ldots,\xi_m \in \R_+$, and rewrite the inequality-constrained problem as one with equality constraints:
\begin{align}
\min_{\theta \in \R^p, \xi\in \R^m_+} \tf(\theta,\lambda) \quad \text{s.t. } \tg(\theta,\lambda) + \xi = \0.
\label{eq:opt-inequality}
\end{align}
We can now apply  Algorithm \ref{algo:implicit} to solve the re-written problem by treating the model parameters and auxiliary variables $(\theta, \xi) \in \R^p \times \mathbb{R}^m_+$ together as the optimization variables, with an additional projection step in the gradient descent procedure to ensure non-negativity of the $\xi_i$s. We did not experiment with this version of the algorithm for the sake of implementation simplicity. 

\remove{
{\bf Fairness constraints.~}
Our approach can also be used to optimize for recent group fairness metrics \cite{zafar2017fairness}. In a typical fairness setup, each example belongs to one of $m$ protected groups, and the goal is to constrain the model to have equitable performance across all groups. One way to enforce this requirement is to introduce a threshold $\lambda_i$ for examples from group $i \in [m]$, and to then tune the $m$ thresholds to satisfy the fairness constraints \cite{hardt2016equality}. This reliance upon thresholds makes these constraints amenable to optimization with implicit functions.
For example, if one wishes to minimize the classification error subject to the 80\% rule \cite{biddle2006adverse}, which requires the proportion of positive predictions to be at least 80\% for each group, one can set this up as an optimization problem of the form \eqref{eq:opt-inequality}, with the objective $f$ defined as the classification error, and the constraints set to:
\[
g_i(\theta,\lambda) \,=\,  \frac{1}{m_i}\left({\tp_i(s^\theta_{\lambda_i}) + \fp_i(s^\theta_{\lambda_i})}\right) - 0.8,
\]
where $\tp_i$ and $\fp_i$ are the true positives and false positives on examples belonging group $i$, and $m_i$ is the total number of examples in group $i$. In practice, one can impose more than one constraint per group, as long as for any $\theta$, one can always find  thresholds $\lambda_i$s for which the constraints are satisfied (which is indeed the case with  popular fairness metrics such as equal opportunity \cite{hardt2016equality}).
}

\subsection{Multi-class metrics}

In our experiments, we handled %
binary classification tasks. The extension to multi-class metrics requires some effort as in this case we are allowed to predict only one among $m$ labels. As with the binary classification setting, 
we work with a model $s^\theta: X \> \R^m$ that outputs $m$ scores, and we will maintain one parameter $\lambda_i$ for each class $i \in [m]$. The parameters  $\lambda_i$'s would then be used to post-shift the model via a \textit{weighted} or \textit{shifted} argmax to predict the final class:
\[
s^\theta_\lambda(x) \in \argmax_{i \in [m]} (s^\theta_i(x) - \lambda_i).
\]
Computing the parameters $\lambda$s so that the resulting classifier satisfies the specified constraints is not straightforward, but can still be performed efficiently with, \eg, the methods in \citet{narasimhan2015consistent}. While it isn't clear if a feasible $\lambda$ exists for general rate constraints, it certainly does for constraints like ``coverage'' \cite{cotter2019optimization}, which require that the model makes a certain percentage of predictions from each class.

\subsection{Ranking metrics}

Perhaps, the most interesting extension of our approach is to query-based ranking problems \cite{schutze2008introduction}, where each example contains a query and a list of documents, and the goal is to rank the documents based on the relevance to the query. Popular ranking metrics such as Precision@$K$ or Recall@$K$ seek to measure performance in the top ranked documents \cite{lapin2017analysis}. Unfortunately, writing these metrics out as an explicit constrained optimization problem would require one constraint per query, with the number of constraints growing with the size of the training set. Consequently, standard constrained optimization approaches, when applied to optimize these metrics, would need to maintain one Lagrange multiplier for each query, making it impractical to use them with large datasets. Recently, \cite{narasimhan2020} propose solving such heavily-constrained problems with lower-dimensional representations of Lagrange multipliers. In contrast, our method offers an alternate route which does not require explicitly handling the large number of constraints, through implicit modeling of the per-query thresholds. We look forward to future work comparing our implicit thresholding approach with the state-of-the-art  methods for these ranking metrics (e.g. \citet{kar2015surrogate,lapin2017analysis}).

%% file: supplementary/appendix.tex
We provide more details, in particular: \\
{\bf Appendix \ref{app:proof}}: Proof of Proposition 1.\\ %
{\bf Appendix \ref{app:timing}}: \textit{Run-time comparisons}, i.e.\ progress in terms of the evaluation metric on the validation and test sets as a function of training time. \\
{\bf Appendix \ref{app:celeba}}: More experiments on CelebA. \\
{\bf Appendix \ref{app:bigearth}}: More experiments on BigEarthNet.

\section{Proof of Proposition 1} %
\label{app:proof}
The assumption in Proposition 1 %
holds, for example,
when  we seek to minimize the FPR subject to FNR $=\beta$. In this case,
the FPR objective can be approximated by $\tilde{f}(\theta, \lambda) = \E_{x \sim \mathcal{D}_0}\left[\ell\left(-1, \theta^\top x + \lambda\right)\right]$ and the FNR constraint can be approximated by
$\tilde{g}(\theta, \lambda) = \E_{x \sim \mathcal{D}_1}\left[\ell\left(+1, \theta^\top x + \lambda\right)\right] - \beta$, where $\ell(y, z) = \log(1+e^{-yz})$ is the standard logistic loss, and $\mathcal{D}_0$ and $\mathcal{D}_1$ are respectively the class-conditional distributions over examples with labels 0 and 1. Note that $\tilde{f}(\theta, \lambda)$ is jointly convex in $(\theta, \lambda)$  and is strictly increasing
in $\lambda$, while $\tilde{g}(\theta, \lambda)$ is jointly convex in $(\theta, \lambda)$  and is strictly decreasing in $\lambda$.
\begin{proof}[Proof of Proposition 1] %
From the joint convexity of $\tg$, we have for any $(\theta, \lambda)$ and $(\theta', \lambda')$:
\[
\langle \nabla_\theta \tg(\theta', \lambda'), \theta - \theta'\rangle + \frac{\partial \tg(\theta', \lambda')}{\partial \lambda} (\lambda - \lambda') \,\leq\, \tg(\theta, \lambda) - \tg(\theta', \lambda') .
\]
Therefore this also holds for $(\theta, \th(\theta))$ and $(\theta', \th(\theta'))$:
\[
\langle \nabla_\theta \tg(\theta', \th(\theta')), \theta - \theta'\rangle + \frac{\partial \tg(\theta', \th(\theta'))}{\partial \lambda} (\th(\theta) - \th(\theta')) \,\leq\, \tg(\theta, \th(\theta)) - \tg(\theta', \th(\theta')) \,=\, 0 - 0 = 0.
\]
Because $g$ is strictly decreasing in $\lambda$, $\frac{\partial \tg(\theta', \th(\theta'))}{\partial \lambda} < 0$, and therefore we can rewrite
the above inequality as:
\[
\frac{1}{\frac{\partial \tg(\theta', \th(\theta'))}{\partial \lambda}}\langle \nabla_\theta \tg(\theta', \th(\theta')), \theta - \theta'\rangle +  \th(\theta) - \th(\theta') \,\geq\, 0,
\]
Using the fact that $\nabla_\theta \th(\theta') = -\frac{ \nabla_\theta \tg(\theta', \th(\theta'))}{\frac{\partial \tg(\theta', \th(\theta'))}{\partial \lambda}}$ (see (4) %
in the main text), we have:
\[
\langle \nabla_\theta \th(\theta'), \theta' - \theta\rangle   \,\geq\, \th(\theta') - \th(\theta),
\]
or 
\[
\th(\theta)  \,\geq\, \th(\theta') + \langle \nabla_\theta \th(\theta'), \theta - \theta'\rangle.
\]
This shows that $\th$ is convex in $\theta$. The convexity of $\tf(\theta, \th(\theta))$ follows from the convexity of 
$\tf$ and $\th$, and from the fact that $\tf$ is monotonically increasing in its second argument.
\end{proof}

\section{Timing comparisons}
\label{app:timing}
\begin{figure}[h]
\centering
\includegraphics[width=0.4\columnwidth]{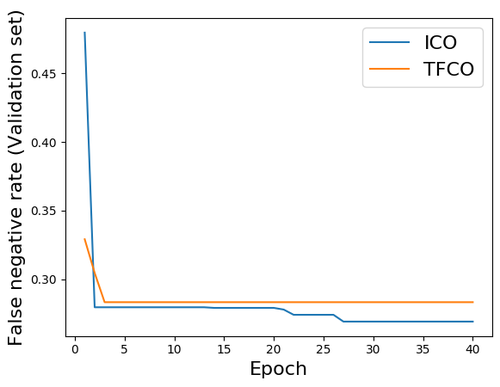}
\includegraphics[width=0.4\columnwidth]{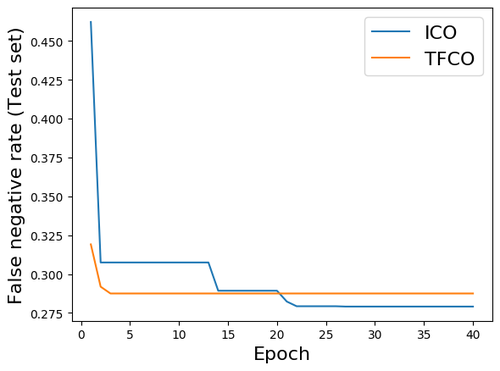}
\caption{\emph{Minimizing false negative rate (FNR) at a fixed false positive rate (FPR) of 0.05 for CelebA}: FNR as a function of training epochs for TFCO and the proposed ICO.}
\label{fig:frr_timing_epochs}
\end{figure}

We monitor the performance of TFCO and ICO in terms of value of the evaluation metric as the training proceeds. At the end of every training epoch, we record the best value of metric seen so far on the validation set, and use the same model (that yields the best validation metric) to score the test set. 

\begin{figure}[t]
\centering
\includegraphics[width=0.4\columnwidth]{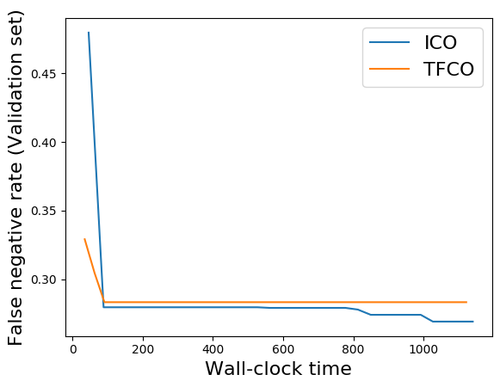}
\includegraphics[width=0.4\columnwidth]{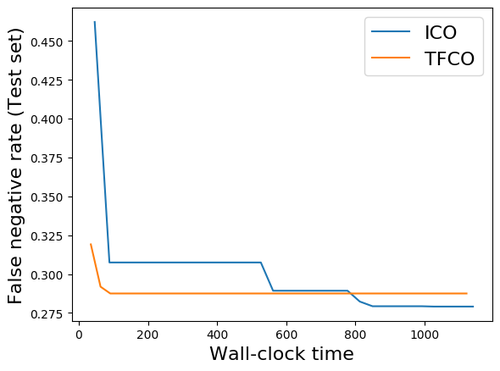}
\caption{\emph{Minimizing false negative rate (FNR) at a fixed false positive rate (FPR) of 0.05 for CelebA}: FNR as a function of wall-clock time for TFCO and the proposed ICO.}
\label{fig:frr_timing_wallclock}
\end{figure}

For the problem of optimizing false negative rate (FNR) at a fixed false positive rate (FPR) on CelebA, Figures \ref{fig:frr_timing_epochs} and \ref{fig:frr_timing_wallclock} show these FNR values for the attribute \emph{High\_Cheekbones} on the validation and test sets as the training proceeds in terms of training epochs and actual wall-clock time, respectively. 
We observe that while TFCO converges to a much lower FNR at the end of the first epoch (the first data point shown in Figure \ref{fig:frr_timing_epochs}), the proposed ICO eventually achieves a lower FNR on both validation and test sets. Both TFCO and ICO were trained for 40 epochs in these experiments and TFCO was about 1.3x faster in terms of wall-clock time.

\begin{figure}[t]
\centering
\includegraphics[width=0.4\columnwidth]{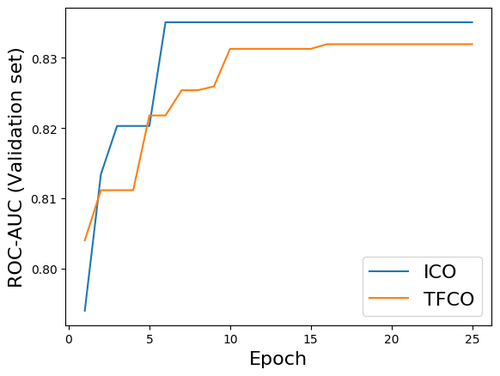}
\includegraphics[width=0.4\columnwidth]{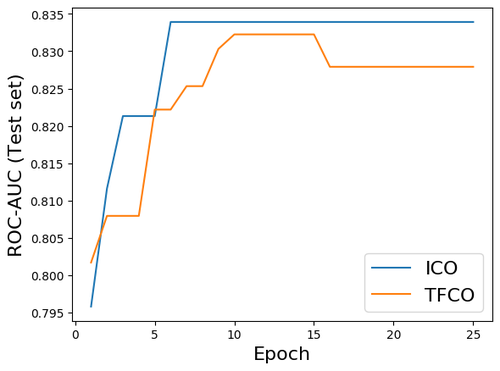}
\caption{\emph{Maximizing partial area under the ROC curve (for FPR $\in[0,0.1]$) for CelebA}: ROC-AUC as a function of training epochs for TFCO and the proposed ICO.}
\label{fig:aucroc_timing_epochs}
\end{figure}

\begin{figure}[t]
\centering
\includegraphics[width=0.4\columnwidth]{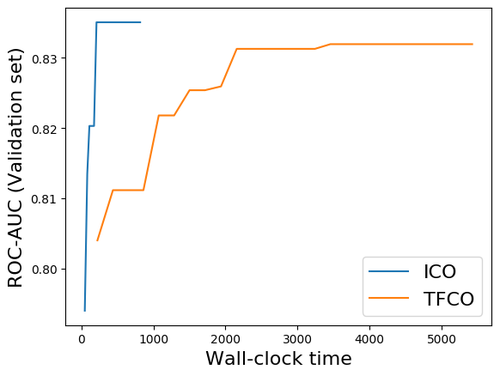}
\includegraphics[width=0.4\columnwidth]{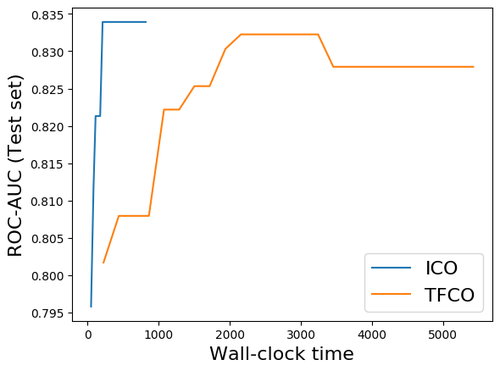}
\caption{\emph{Maximizing partial area under the ROC curve (for FPR $\in[0,0.1]$) for CelebA}: ROC-AUC as a function of wall-clock time for TFCO and the proposed ICO.}
\label{fig:aucroc_timing_wallclock}
\end{figure}

\remove{
\begin{figure}[t]
\centering
\includegraphics[width=0.4\columnwidth]{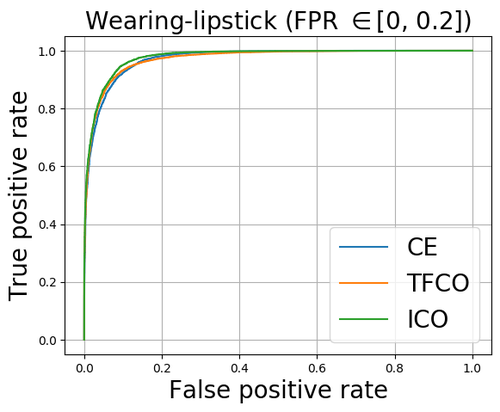}
\includegraphics[width=0.4\columnwidth]{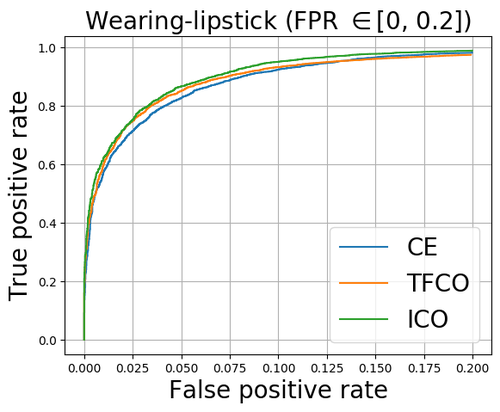}\\
\includegraphics[width=0.4\columnwidth]{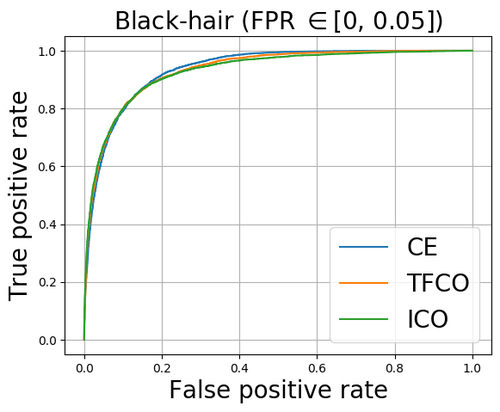}
\includegraphics[width=0.4\columnwidth]{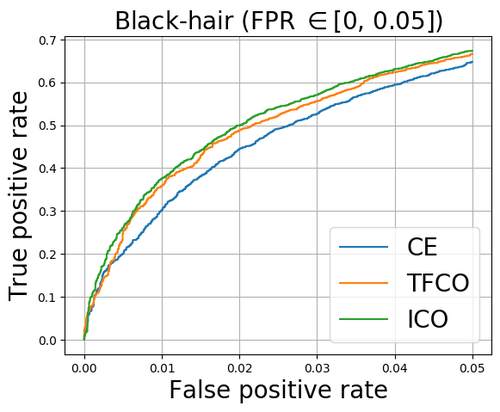}
\caption{\emph{ROC curves for CelebA}: {\bf Top:} For attribute \emph{Wearing-lipstick} and optimizing for ROC-AUC with FPR $\in [0,0.2]$, {\bf Bottom:} For attribute \emph{Black-hair} and optimizing for ROC-AUC with FPR $\in [0,0.05]$. Left figures show full ROC curves while the right figures show the ROC curves in the respective target FPR ranges.}
\label{fig:celeba_roc_plots}
\end{figure}
}

We repeat similar experiment for the problem of optimizing the partial area under the ROC curve for FPR in the range of $[0,0.1]$, again for CelebA. Figure \ref{fig:aucroc_timing_epochs} and 
\ref{fig:aucroc_timing_wallclock} show the ROC-AUC on validation and test sets for the \emph{High\_Cheekbones} attributes as the training proceeds in terms of epochs and wall-clock time, respectively. We observe similar behavior as in the earlier experiment and TFCO converges to a much better ROC-AUC early on in the training at the end of first epoch (first data point in the plots). However, the proposed ICO eventually achieves a better ROC-AUC both on validation and test sets. Both TFCO and ICO were trained for 25 epochs in this experiment and ICO is about 5x faster than TFCO in terms of wall-clock time. This is due to the fact that optimizing ROC-AUC is a problem with multiple constraints (10 in this case) and we do not optimize the thresholds using gradients in ICO and only rely on the threshold correction step after every 100 minibatches. On the other hand, TFCO training time per minibatch slows down due to multiple constraints.

\begin{table*}[b]
	\centering
	\caption{Minimizing false negative rate (FNR) at a given false positive rate (FPR) for {\bf CelebA}.~~ The mean FNR (in \%) are reported over five random trials, along with std. deviations, for cross-entropy (CE), TFCO and \proposed. Proposed \proposed~outperforms both CE and TFCO by a considerable margin in most cases. \textit{Lower} values are better. We also list the hyperparameters for ICO: surrogate function for both objective and constraint is taken to be softplus; correction step in Alg.~\ref{algo:implicit} is applied every 1000 minibatch steps ($N=1000$) using data from next 10 minibatches ($k=10$); the temperature hyperparameter $\tau$ for softplus is provided in the Table below.}
	\vspace{2mm}
	\begin{tabular}{c c c c c| c}
	\toprule[0.25ex]
	 Attributes & FPR & CE & TFCO & ICO &  ICO hyperparameter $\tau$ \\[1mm]
	 \midrule
	 \midrule
    \multirow{4}{*}{High-cheekbones} & 1\% & 53.52 (1.71)& 49.09 (1.56)& {\bf 46.96 (0.55)}  & 0.01 \\
     & 2\% & 44.82 (1.23)& 40.89 (0.82)& {\bf 39.84 (0.49)} & 0.01\\
     & 5\% & 32.87 (0.83)& 30.11 (0.67)& {\bf 28.54 (0.30)}& 0.01\\
     & 10\% & 22.88 (0.43)& 20.37 (0.65)& {\bf 19.66 (0.36)}& 0.01\\
    \midrule

    \multirow{4}{*}{Heavy-makeup} & 1\% & 57.00 (0.84)& 52.07 (1.24)& {\bf 49.65 (0.81)}& 1 \\
    & 2\% & 45.59 (1.24)& 41.23 (1.41)& {\bf 38.89 (0.80)} & 0.001\\
     & 5\% & 28.22 (0.80)& 25.43 (0.97)& {\bf 23.11 (0.70)} & 0.001\\
     & 10\% & 15.12 (0.44)& 13.61 (0.77)& {\bf 12.36 (0.19)} & 0.01\\
     \midrule

    \multirow{4}{*}{Wearing-lipstick} & 1\% & 44.00 (1.28)& 42.64 (1.29) & {\bf 37.47 (0.97)} & 1\\
    & 2\% & 32.74 (0.88)& 30.44 (1.51)& {\bf 26.74 (0.50)} & 0.01\\
     & 5\% & 16.33 (0.43)& 14.97 (0.77)& {\bf 13.05 (0.25)} & 0.001\\
     & 10\% & 6.61 (0.27)& 5.92 (0.21)& {\bf 4.78 (0.12)} & 0.001\\
     \midrule
    
    \multirow{4}{*}{Smiling} & 1\% &37.40 (1.30) & 35.93 (1.37)& {\bf 33.74 (0.71)} & 0.01 \\
    & 2\% & 29.44 (0.76)& 27.80 (1.23)& {\bf 26.10 (0.57)} & 0.01\\
     & 5\% & 18.73 (0.53)& 17.04 (0.80)& {\bf 16.88 (0.25)} & 0.01\\
     & 10\% & 11.78 (0.20)& 10.74 (0.29)& {\bf 10.23 (0.25)} & 0.01\\
     \midrule
    
    \multirow{4}{*}{Black-hair} & 1\% & 69.32 (1.78) & 64.47 (1.55) & {\bf 63.23 (1.19)} & 0.001 \\
    & 2\%  & 56.48 (1.48) & 52.00 (1.10) & {\bf 50.50 (0.67)} & 0.001\\
    & 5\% & 36.72 (1.61) & {\bf 32.41 (0.60)} & 32.48 (0.66) & 0.001 \\
    & 10\% & 22.97 (1.87) & 19.16 (1.22) & {\bf 18.62 (0.43)} & 0.001 \\
    \midrule
    Blond-hair & 1\% & 40.49 (1.18)& 38.62 (1.17)& {\bf 36.85 (0.58)} & 0.01\\
    & 2\% & 28.89 (1.17)& 25.64 (1.20)& {\bf 24.20 (0.72)} & 0.001\\
     & 5\% & 13.44 (1.01) & 11.64 (0.76)& {\bf 10.81 (0.24)} & 0.001\\
     & 10\% & 6.54 (0.46)& 4.91 (0.22)& {\bf 4.68 (0.20)} & 0.001\\
     \midrule
    
    Brown-hair & 1\% & 80.75 (1.82)& 77.16 (0.74)& {\bf 76.34 (0.78)} & 0.001 \\
    & 2\% & 69.69 (1.77)& 66.10 (1.04)& {\bf 65.74 (1.28)} & 0.001\\
     & 5\% & 52.41 (2.55)& {\bf 45.83 (0.92)}& 46.43 (0.51) & 0.001\\
     & 10\% & 35.92 (2.71)& {\bf 29.94 (0.87)} & 30.02 (0.67)& 0.001\\
     \midrule
    
    Wavy-hair & 1\% & 85.04 (0.80)& 84.42 (0.95)& {\bf 83.54 (0.69)} & 0.001 \\
    & 2\% & 78.91 (1.20)& 77.02 (1.47)& {\bf 76.07 (0.90)} & 0.001\\
     & 5\% & 65.79 (1.77)& 61.81 (0.92)& {\bf 60.71 (1.01)} & 0.001\\
     & 10\% & 50.52 (1.41)& 47.49 (0.98)& {\bf 45.95 (1.12)} & 0.001\\
	\bottomrule[0.25ex]
	\end{tabular}
	\label{tab:celeba_frr_app}
\end{table*}

\begin{table*}[b]
	\centering
	\caption{Maximizing area under the ROC curve for {\bf CelebA}, in a given FPR range $[0,\beta]$ for $\beta\in\{1\%,2\%,5\%,10\%,20\%\}$.~~ The mean ROC-AUC are reported over five random trials, along with std. deviations, for cross-entropy (CE), Pairwise-loss, TFCO and \proposed.  \textit{Higher} values are better. We also list the hyperparameters for ICO: surrogate function for both objective and constraint is taken to be sigmoid; the correction step in Alg.~\ref{algo:implicit} is applied every $N$ minibatch steps using data from next $k$ minibatches. The values of $N$, $k$, and the temperature hyperparameter $\tau$ for sigmoid are selected on the validation set and are provided in the Table below.}
	\vspace{2mm}
	\begin{tabular}{c c c c c c| c}
	\toprule[0.25ex]
	 Attributes & FPR & CE & Pairwise-loss & TFCO & ICO & ICO hyperparameters \\[1mm]
	 \midrule
	 \midrule
    \multirow{4}{*}{High-cheekbones} & 1\% & 66.10 (1.14)& 68.18 (2.01) & 62.96 (10.45)& {\bf 69.83 (0.84)} &  $\tau=1,N=1000,k=100$  \\
     & 2\% & 70.87 (0.91)& 72.85 (0.28) & {\bf 74.98 (0.20)}& 73.17 (0.84) & $\tau=1,N=1000,k=100$\\
     & 5\% &75.89 (1.26) & 78.13 (0.36) & 73.83 (11.30)& {\bf 78.45 (0.53)} & $\tau=1,N=1000,k=100$\\
     & 10\% & 80.15 (1.02)& 81.51 (0.57)& 74.07 (12.13)& {\bf 82.67 (0.49)} & $\tau=0.1,N=100,k=100$\\
     & 20\% & 84.26 (0.74)& 85.70 (0.33) & 72.44 (17.48)& {\bf 86.82 (0.30)} & $\tau=0.1,N=100,k=100$\\
    \midrule

    \multirow{4}{*}{Heavy-makeup} & 1\% & 65.03 (0.92) & 66.33 (1.41) & {\bf 68.55 (0.83)} & 66.75 (0.31) & $\tau=1,N=1000,k=100$  \\
     & 2\% & 68.82 (0.61)& 71.13 (0.80) & {\bf 73.43 (0.13)} & 71.57 (0.80) & $\tau=1,N=1000,k=100$\\
     & 5\% & 75.48 (1.44)& 77.78 (0.40) & {\bf 79.54 (0.09)}& 78.32 (0.33) & $\tau=0.1,N=100,k=100$\\
     & 10\% & 81.53 (1.10)& 82.85 (0.75) & {\bf 85.00 (0.12)}& 84.39 (0.47) & $\tau=0.1,N=100,k=100$\\
     & 20\% & 88.22 (0.40)& 88.68 (0.36) & 89.93 (0.11)& 89.81 (0.19) & $\tau=0.1,N=100,k=100$\\
     \midrule

    \multirow{4}{*}{Wearing-lipstick} & 1\% & 70.24 (1.13)& 71.77 (0.70) & {\bf 74.82 (0.33)}& 72.29 (0.77) & $\tau=0.01,N=100,k=100$   \\
     & 2\% & 75.42 (0.68)& 77.21 (0.39) & {\bf 79.28 (0.22)}& 78.44 (0.28) & $\tau=1,N=1000,k=100$\\
     & 5\% & 82.19 (0.37)& 83.42 (0.97) & 84.68 (0.19)& 84.56 (0.28) & $\tau=0.01,N=100,k=100$\\
     & 10\% & 87.70 (0.89)& 88.44 (0.25) & 89.35 (0.18) & {\bf 89.73 (0.27)} & $\tau=0.01,N=100,k=100$\\
     & 20\% & 91.88 (0.44) & 93.00 (0.20) & 93.19 (0.12) & {\bf 93.93 (0.15)} & $\tau=0.1,N=100,k=100$\\
     \midrule
    
    \multirow{4}{*}{Smiling} & 1\% & 75.39 (0.51)& 75.87 (0.63) & {\bf 78.03 (0.42)}& 75.59 (0.76) & $\tau=1,N=1000,k=100$   \\
     & 2\% & 78.44 (0.41)& 79.85 (0.52) & {\bf 81.51 (0.20)}& 79.80 (0.55) & $\tau=1,N=1000,k=100$\\
     & 5\% & 83.48 (0.68)& 84.50 (0.54) & 64.97 (17.24)& {\bf 84.81 (0.41)} & $\tau=1,N=1000,k=100$\\
     & 10\% & 86.76 (0.84)& 88.06 (0.22) & 73.79 (18.63)& {\bf 88.80 (0.30)} & $\tau=0.1,N=100,k=100$\\
     & 20\% & 90.88 (0.52)& 91.46 (0.11) & 76.61 (18.69)& {\bf 92.08 (0.09)} & $\tau=0.1,N=100,k=100$\\
     \midrule
    
    \multirow{4}{*}{Black-hair} & 1\% & 60.53 (0.86)& 57.73 (1.11) & {\bf 61.44 (0.44)} & 61.24 (0.26) & $\tau=1,N=1000,k=100$  \\
     & 2\% & 64.48 (0.45)& 63.93 (1.10) & {\bf 66.53 (0.44)} & 66.07 (0.61) & $\tau=1,N=1000,k=100$\\
     & 5\% & 70.87 (1.01)& 71.85 (0.25) & 73.19 (0.29)& {\bf 73.79 (0.45)} & $\tau=0.1,N=1000,k=100$\\
     & 10\% & 78.04 (0.71)& 78.49 (0.56) & 79.88 (0.11)& {\bf 80.19 (0.07)} & $\tau=0.1,N=100,k=100$\\
     & 20\% & 84.33 (0.54)& 85.45 (0.37) & 86.00 (0.09)& 86.09 (0.30) & $\tau=0.1,N=100,k=100$\\
    \midrule
    Blond-hair & 1\% & 71.36 (0.62)& 70.38 (0.89) &  {\bf 73.11 (0.46)}& 72.11 (0.37)   & $\tau=0.1,N=1000,k=100$\\
     & 2\% & 76.50 (0.91)& 76.25 (0.66) & {\bf 79.01 (0.16)} &78.06 (0.54) & $\tau=0.1,N=1000,k=100$\\
     & 5\% & 84.74 (0.69)& 84.21 (0.37) & {\bf 86.18 (0.13)}& 85.76 (0.25) & $\tau=0.1,N=1000,k=100$\\
     & 10\% & 89.39 (0.59)& 89.75 (0.73)& {\bf 90.63 (0.15)}& 90.49 (0.21) & $\tau=0.1,N=1000,k=100$\\
     & 20\% & 93.30 (0.33)& 93.56 (0.38)& 94.24 (0.10)& 94.27 (0.14) & $\tau=0.1,N=100,k=100$\\
     \midrule
    
    Brown-hair & 1\% & 55.40 (0.54)& 52.65 (0.56) &  56.09 (0.28)& {\bf 56.61 (0.41)}    & $\tau=1,N=1000,k=100$\\
     & 2\% & 58.82 (0.38)& 54.62 (1.34)& 59.68 (0.21)& {\bf 60.10 (0.40)} & $\tau=1,N=1000,k=100$\\
     & 5\% &64.87 (0.63)& 61.21 (2.25)& 66.71 (0.11)& {\bf 67.13 (0.40)} & $\tau=1,N=1000,k=100$\\
     & 10\% & 69.74 (1.14)& 70.05 (0.34)& {\bf 73.23 (0.27)}& 73.01 (0.25) & $\tau=1,N=1000,k=100$\\
     & 20\% & 77.67 (1.03)& 78.25 (0.51)& 80.09 (0.16)& 80.06 (0.23) & $\tau=0.1,N=1000,k=100$\\
     \midrule
    
    Wavy-hair & 1\% & 54.03 (0.28)& 50.91 (0.23) &  52.36 (1.31)& {\bf 54.33 (0.09)}   & $\tau=1,N=1000,k=100$ \\
     & 2\% & 55.85 (0.78)& 52.08 (0.50)& 52.64 (2.14)& {\bf 57.02 (0.31)} & $\tau=1,N=1000,k=100$\\
     & 5\% & 60.16 (0.73)& 54.17 (2.30)& 53.70 (2.98)& {\bf 61.30 (0.38)} & $\tau=1,N=1000,k=100$\\
     & 10\% & 64.42 (0.37)& 56.70 (2.09)& 57.16 (4.32)& {\bf 65.34 (0.48)} & $\tau=0.1,N=100,k=100$\\ 
     & 20\% & 69.26 (1.03)& 68.56 (0.49)& 66.47 (5.60)& {\bf 71.48 (0.32)} & $\tau=0.1,N=100,k=100$\\
	\bottomrule[0.25ex]
	\end{tabular}
	\label{tab:celeba_auc_app}
\end{table*}

\section{CelebA results}
\label{app:celeba}
We report results on more CelebA attributes for the two problems considered in the main text: (i) Minimizing false negative rate (FNR) at a fixed false positive rate (FPR) for FPRs $\in \{1\%,2\%,5\%,10\%\}$ (Table \ref{tab:celeba_frr_app}), and (ii) Maximizing partial area under the ROC curve (ROC-AUC) for FPR in the range $[0,\beta]$ for $\beta\in\{1\%,2\%,5\%,10\%,20\%\}$ (Table \ref{tab:celeba_auc_app}). These results also show the standard deviation over five random trials which were omitted in the main text due to space constraints. For partial AUC in the FPR range $[0,\beta]$, we also compare with a pairwise loss baseline \cite{narasimhan2013svmpauctight} which optimizes the objective $\frac{1}{N^+|S^-|}\sum_{i:y_i=1} \sum_{j\in S^{-}} \tilde{f}(s^\theta(x_i)-s^\theta(x_j))$, where $s^\theta(x)$ denotes the score (\eg, logits) for example $x$, $N^+$ is the number of positive examples in the minibatch, $S^-$ is the subset of negative examples whose scores lie in the top $\beta$ fraction of all negative examples, and $\tilde{f}$ is the surrogate used for 0-1 loss (either softplus or sigmoid with a temperature hyperparameter as we used for the proposed method). %

\begin{table*}[b]
	\centering
	\caption{Maximizing area under the ROC curve for {\bf BigEarthNet}, in a given FPR range $[0,\beta]$ for $\beta\in\{5\%,10\%,20\%\}$.~~ The mean ROC-AUC are reported over five random trials, along with std. deviations, for cross-entropy, Pairwise-loss, TFCO, \proposed. \textit{Higher} values are better.}
	\vspace{2mm}
	\begin{tabular}{c c c c c c}
	\toprule[0.25ex]
	 Labels & FPR & CE & Pairwise-loss & TFCO & ICO  \\[1mm]
	 \midrule
	 \midrule
    \multirow{3}{*}{Broad-Leaved Forest (BLF)} & 5\% & 66.20 (0.59)& 52.07 (0.35) & 66.43 (1.86)& {\bf 69.90 (0.53)}\\
    & 10\% & 71.00 (0.80)& 53.78 (1.34) & 71.72 (0.78)& {\bf 73.91 (0.66)}\\
    & 20\% &75.42 (0.67) & 57.94 (0.81) & 76.20 (0.61) & {\bf 77.91 (0.77)}\\
    \midrule

    \multirow{3}{*}{Complex Cultivation patterns (CC)} & 5\% & 62.19 (0.48)& 52.44 (0.26)& 62.71 (2.19)& {\bf 63.61 (0.12)}\\
    & 10\% & 67.46 (0.25)& 54.71 (0.62) & 66.75 (0.82)& {\bf 68.35 (0.42)} \\
    & 20\% & 73.81 (0.83)& 59.34 (0.64)& {\bf 76.01 (0.77)}& 74.88 (0.43) \\
    \midrule

    \multirow{3}{*}{Coniferous Forest (CF)} & 5\% & 71.93 (0.66)& 59.00 (0.56)& 71.49 (3.44)& {\bf 74.70 (0.62})\\
    & 10\% & 78.62 (0.59)& 77.66 (2.22)& 79.98 (1.26)& {\bf 80.76 (0.94)}\\
    & 20\% & 84.82 (0.68)& 85.84 (0.31)& {\bf 86.62 (0.62)}& 86.02 (0.29) \\
    \midrule

    \multirow{3}{*}{Discontinuous Urban Fabric (DUF)} & 5\% & 69.80 (1.45)& 55.33 (0.98) & 71.76 (1.89)& {\bf 73.94 (0.39)}\\
    & 10\% & 75.13 (0.86)& 59.15 (2.17) & 77.20 (0.87)& {\bf 78.03 (0.32)}\\
    & 20\% & 78.86 (1.19)& 78.89 (0.38)& 81.45 (0.62)& {\bf 81.83 (0.57)} \\
    \midrule

    \multirow{3}{*}{\parbox{5cm}{Land principally occupied by Agriculture,
with significant areas of Natural Vegetation (ANV)}} & 5\% & 58.79 (0.49)& 51.23 (0.16) & 58.80 (0.77)& {\bf 60.46 (0.17)}\\
    & 10\% & 62.72 (0.46)& 52.65 (0.56) & 63.89 (0.98)& {\bf 64.38 (0.32)} \\
    & 20\% & 67.77 (0.34)& 54.36 (0.63) & 69.17 (0.52)& 68.97 (0.76)\\
    \midrule

    \multirow{3}{*}{Mixed Forest (MF)} & 5\% & 64.48 (0.76)& 54.59 (0.60) & 65.50 (0.30)& {\bf 65.93 (0.60)}\\
    & 10\% & 71.05 (0.47)& 60.41 (0.28) & 72.06 (0.33)& 71.89 (0.48)\\
    & 20\% & 77.56 (0.69)& 76.44 (0.26) & 78.83 (0.14)& {\bf 79.20 (0.26)}\\
    \midrule

    \multirow{3}{*}{Non-Irrigated Arable Land (NIAL} & 5\% & 70.07 (0.27)& 55.10 (0.28) & {\bf 72.67 (0.19)} &71.45 (0.48) \\
    & 10\% & 75.29 (0.37)& 59.62 (2.37) & {\bf 77.30 (0.12)}& 76.78 (0.76)\\
    & 20\% &79.97 (0.67) & 80.23 (0.17) &{\bf 82.05 (0.09)} &81.72 (0.56) \\
    \midrule

    \multirow{3}{*}{Pastures} & 5\% & 72.70 (0.46)& 59.41 (0.85) & {\bf 74.16 (0.29)}& 73.61 (0.55)\\
    & 10\% & 75.95 (0.55)& 74.78 (0.29)& 78.04 (0.31)& 77.85 (0.65)\\
    & 20\% & 80.31 (0.87)&80.42 (0.66) & {\bf 82.38 (0.17)}& 82.10 (0.18)\\
    \midrule

    \multirow{3}{*}{Transitional Woodland/Shrub (TWS)} & 5\% &57.12 (0.32) & 51.30 (0.32) &58.21 (0.08) &{\bf 59.64 (0.27)} \\
    & 10\% & 60.24 (0.21)& 52.45 (0.61)& 61.82 (0.13)& {\bf 62.82 (0.61)}\\
    & 20\% & 64.98 (0.92)& 55.47 (0.31)& 67.15 (0.26)& 67.24 (1.10)\\
    \midrule

    \multirow{3}{*}{Water Bodies (WB)} & 5\% & 76.52 (0.59)& 54.29 (1.35)& 77.47 (0.48)&{\bf 78.71 (0.33)} \\
    & 10\% & 80.81 (0.69)& 57.23 (0.73)& 81.76 (0.34)& {\bf 82.79 (0.35)}\\
    & 20\% & 85.27 (0.39)& 86.11 (0.25)& 85.46 (0.19)& {\bf 86.66 (0.31)}\\
	\bottomrule[0.25ex]
	\end{tabular}
	\label{tab:bigearth_auc_app}
\end{table*}

\begin{table*}[b]
	\centering
	\caption{Maximizing area under the ROC curve for {\bf BigEarthNet}, in a given FPR range $[0,\beta]$ for $\beta\in\{5\%,10\%,20\%\}$:~ hyperparameters for ICO selected using the validation set.~~Surrogate function for both objective and constraint is taken to be softplus; the correction step in Alg.~\ref{algo:implicit} is applied every $N$ minibatch steps using data from next $k$ minibatches. The values of $N$, $k$, and the temperature hyperparameter $\tau$ for sigmoid are selected on the validation set and are provided in the Table below.}
	\vspace{2mm}
	\begin{tabular}{c c c}
	\toprule[0.25ex]
	 Labels & FPR & ICO hyperparameters \\[1mm]
	 \midrule
	 \midrule
    \multirow{3}{*}{Broad-Leaved Forest (BLF)} & 5\% & $\tau=5,N=1000,k=50$\\
    & 10\% &  $\tau=5,N=1000,k=50$\\
    & 20\% & $\tau=5,N=1000,k=50$\\
    \midrule

    \multirow{3}{*}{Complex Cultivation patterns (CC)} & 5\% & $\tau=1,N=100,k=50$\\
    & 10\% &  $\tau=0.001,N=100,k=10$\\
    & 20\% &  $\tau=0.001,N=100,k=10$\\
    \midrule

    \multirow{3}{*}{Coniferous Forest (CF)} & 5\% & $\tau=0.001,N=100,k=10$\\
    & 10\% & $\tau=5,N=1000,k=100$\\
    & 20\% &  $\tau=0.001,N=100,k=50$\\
    \midrule

    \multirow{3}{*}{Discontinuous Urban Fabric (DUF)} & 5\% & $\tau=5,N=1000,k=100$\\
    & 10\% & $\tau=5,N=1000,k=50$\\
    & 20\% &  $\tau=5,N=1000,k=100$\\
    \midrule

    \multirow{3}{*}{\parbox{5cm}{Land principally occupied by Agriculture,
with significant areas of Natural Vegetation (ANV)}} & 5\% & $\tau=1,N=100,k=50$\\
    & 10\% & $\tau=5,N=1000,k=50$ \\
    & 20\% & $\tau=5,N=1000,k=10$\\
    \midrule

    \multirow{3}{*}{Mixed Forest (MF)} & 5\% & $\tau=5,N=1000,k=100$\\
    & 10\% & $\tau=5,N=1000,k=10$\\
    & 20\% & $\tau=5,N=1000,k=10$\\
    \midrule

    \multirow{3}{*}{Non-Irrigated Arable Land (NIAL} & 5\% & $\tau=1,N=100,k=50$ \\
    & 10\% & $\tau=5,N=1000,k=50$\\
    & 20\% & $\tau=5,N=1000,k=50$\\
    \midrule

    \multirow{3}{*}{Pastures} & 5\% & $\tau=1,N=1000,k=50$\\
    & 10\% & $\tau=5,N=1000,k=50$\\
    & 20\% & $\tau=5,N=1000,k=50$\\
    \midrule

    \multirow{3}{*}{Transitional Woodland/Shrub (TWS)} & 5\% & $\tau=5,N=100,k=50$\\
    & 10\% & $\tau=5,N=100,k=50$\\
    & 20\% & $\tau=5,N=100,k=50$\\
    \midrule

    \multirow{3}{*}{Water Bodies (WB)} & 5\% &  $\tau=5,N=1000,k=50$\\
    & 10\% & $\tau=5,N=1000,k=100$\\
    & 20\% & $\tau=5,N=1000,k=50$\\
	\bottomrule[0.25ex]
	\end{tabular}
	\label{tab:bigearth_auc_hyper}
\end{table*}
\section{BigEarthNet results}
\label{app:bigearth}
We also report results on more BigEarthNet labels for the problem of  maximizing partial area under the ROC curve (ROC-AUC) for FPR in the range $[0,\beta]$ for $\beta\in\{5\%,10\%,20\%\}$ (Table \ref{tab:bigearth_auc_app}). These results also show the standard deviation over five random trials which were omitted in the main text due to space constraints. 
We also compare with the pairwise loss baseline for partial AUC as described earlier. 
The proposed \proposed~outperforms  cross-entropy and pairwise loss baselines in all the cases, and also outperforms TFCO for most cases.